\PassOptionsToPackage{dvipsnames}{xcolor}
\documentclass{article}
\usepackage[final]{colm2026_conference}

\usepackage{microtype}
\usepackage{hyperref}
\usepackage{url}
\usepackage{booktabs}
\usepackage{graphicx}
\usepackage{amsmath}
\usepackage{enumitem}
\usepackage{lineno}
\usepackage{xcolor}
\usepackage{longtable}
\usepackage{booktabs, siunitx, caption}
\usepackage{float}
\usepackage{amssymb}
\usepackage{tcolorbox}
\tcbuselibrary{skins}

\usepackage{graphicx}
\usepackage{tikz}
\usetikzlibrary{positioning, arrows.meta, calc, arrows}

\definecolor{darkblue}{rgb}{0, 0, 0.5}
\hypersetup{colorlinks=true, citecolor=darkblue, linkcolor=darkblue, urlcolor=darkblue}

\title{\textbf{BidirLM}: From Text to Omnimodal Bidirectional Encoders by Adapting and Composing Causal LLMs}

\storeabstract{Transforming causal generative language models into bidirectional encoders offers a powerful alternative to BERT-style architectures. However, current approaches remain limited: they lack consensus on optimal training objectives, suffer from catastrophic forgetting at scale, and fail to flexibly integrate the vast ecosystem of specialized generative models. In this work, through systematic ablations on the Gemma3 and Qwen3 families, we identify the key factors driving successful adaptation, highlighting the critical role of an often-omitted prior masking phase. To scale this process without original pre-training data, we introduce a dual strategy combining linear weight merging with a lightweight multi-domain data mixture that mitigates catastrophic forgetting. Finally, we augment our encoders by merging them with specialized causal models, seamlessly transferring modality- and domain-specific capabilities. This open-source recipe, designed for any causal decoder LLM, yields \textit{BidirLM}, a family of five encoders that outperform alternatives on text, vision, and audio representation benchmarks.}

\metadata{Correspondence}{\href{mailto:nicolas.boizard@centralesupelec.fr}{\textcolor{BlueViolet}{nicolas.boizard@centralesupelec.fr}}}
\metadata{Website}{\href{https://huggingface.co/BidirLM}{\textcolor{BlueViolet}{\nolinkurl{https://huggingface.co/BidirLM}}}}
\metadata{Date}{April 2, 2026}

\author{Nicolas Boizard\textsuperscript{1,3} \quad
Théo Deschamps-Berger\textsuperscript{1} \\ [0.2em]
\textmd{Hippolyte Gisserot-Boukhlef\textsuperscript{2,3} \quad
Céline Hudelot\textsuperscript{3}} \quad
Pierre Colombo\textsuperscript{4}\\[0.6em]
\textsuperscript{1}Diabolocom \quad
\textsuperscript{2}Artefact Research Center \\[0.2em] 
\textsuperscript{3}MICS, CentraleSupélec, Université Paris-Saclay \quad
\textsuperscript{4}Cohere
}

\newcommand{\keyfinding}[1]{%
    \begin{tcolorbox}[
        colback=blue!5, colframe=black, boxrule=0.5pt, arc=4pt, boxsep=0pt,
        left=4pt, right=4pt, top=3pt, bottom=3pt,
        before=\vspace{3pt}\noindent, after=\vspace{0pt}
    ]
    \small \textbf{Key finding:} #1
    \end{tcolorbox}%
}

\begin{document}

\ifcolmsubmission
\linenumbers
\fi

\vspace*{-1.4cm}
\maketitle

\section{Introduction}
Causal large language models (LLMs) are not only dominant as generators but serve as the foundation of a vast ecosystem of specialized variants: code~\citep{lozhkov2024starcoder2stackv2}, mathematics~\citep{shao2024deepseekmathpushinglimitsmathematical}, safety~\citep{zhao2025qwen3guardtechnicalreport}, vision~\citep{bai2025qwen3vltechnicalreport}, and audio~\citep{shi2026qwen3asrtechnicalreport}, collectively representing millions of GPU hours of open-source knowledge. Yet, representation tasks remain bound to bidirectional encoders~\citep{devlin2019bertpretrainingdeepbidirectional, he2023debertav3improvingdebertausing, boizard2025eurobertscalingmultilingualencoders}, leaving this knowledge untapped. Repurposing causal models into encoders is therefore a compelling goal, and recent work has begun to explore this direction~\citep{ma2023finetuningllamamultistagetext, behnamghader2024llm2veclargelanguagemodels, wang2024improvingtextembeddingslarge, babakhin2025llamaembednemotron8buniversaltextembedding, gisserotboukhlef2026pretrainencodersmaskedlanguage}. However, this adaptation landscape remains fragmented around three core questions. 
We address each through a fully open-source framework, validated under strictly identical conditions on Gemma3 and Qwen3.

\begin{enumerate}[leftmargin=0pt, topsep=0pt, label={}]
    \item \textbf{What drives adaptation quality?} Existing methods conflate critical design choices like training objectives and attention mechanisms, leaving no consensus on what drives quality. Through controlled ablations (\autoref{sec:experimental_setup}, \autoref{sec:Adaptation_Strategies}), we disentangle these factors and show that enabling bidirectional attention via a masking objective (a step often omitted) is critical to unlock performance on task-specific benchmarks, while contrastive objectives primarily drive generic embedding quality.
    
    \item \textbf{Can adaptation scale without the original pre-training data?} Many adapted models are developed by the same organizations that trained the underlying base models~\citep{vera2025embeddinggemmapowerfullightweighttext,zhang2025qwen3embeddingadvancingtext}. This raises concerns about reproducibility, as these adaptations may implicitly benefit from alignment with undisclosed pre-training corpora, potentially masking the catastrophic forgetting that occurs under distribution shifts. To scale adaptation under strict independent data constraints (\autoref{sec:scaling}), we propose a dual strategy combining training-free linear weight merging with a lightweight multi-domain data mixture. This approach yields adapted models that outperform current open-source alternatives (\autoref{sec:frontier_performance}).
    
    \item \textbf{Can adapted encoders compose with the causal ecosystem?} Current adapted encoders rely on rigid pipelines that fail to compose with other specialized causal models derived from the same backbone. By ignoring the vast ecosystem that motivates starting from causal architectures, these methods leave thousands of GPU hours of open-source specialization unused. We address this by pushing the boundaries of weight merging (\autoref{sec:alignment_and_domain_specificity}). We seamlessly integrate knowledge from specialized causal variants, extending our encoders to new domains and modalities (safety, audio, vision) without requiring full pipeline re-training.
\end{enumerate}

Accompanying this work, we release the multilingual \textit{BidirLM} series, which outperforms open-source alternatives on text, vision, and audio representation benchmarks: BidirLM-270M/1B (Gemma3-based), BidirLM-0.6B/1.7B (Qwen3-based), and BidirLM-Omni-2.5B (text, vision, audio), alongside our training corpus, checkpoints, and experimental variants.

\section{Experimental Setup}\label{sec:experimental_setup}
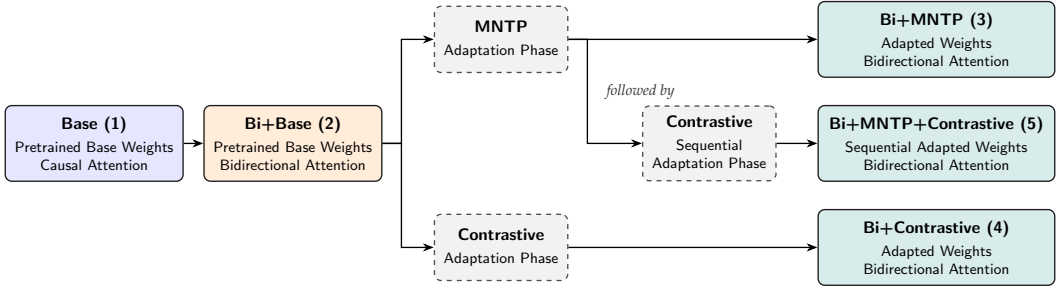
\begin{figure}[h]
\centering
\resizebox{\linewidth}{!}{%
\begin{tikzpicture}[
    >=Stealth,
    font=\sffamily,
    thick,
    model/.style={rectangle, draw=black!90, fill=white, thick, align=center, rounded corners, minimum height=1.6cm, inner sep=2mm},
    final model/.style={model, minimum width=5cm}, 
    process/.style={rectangle, draw=black!70, fill=gray!10, thick, dashed, align=center, rounded corners, minimum height=1.4cm, inner sep=2mm}
]

\node[model, fill=blue!10] (base) at (0, 0) {\textbf{Base (1)}\\[1mm]\footnotesize Pretrained Base Weights\\\footnotesize Causal Attention};

\node[model, fill=orange!15] (bibase) at (4.2, 0) {\textbf{Bi+Base (2)}\\[1mm]\footnotesize Pretrained Base Weights\\\footnotesize Bidirectional Attention};

\node[process] (recon) at (8.6, 2.2) {\textbf{MNTP}\\[1mm]\footnotesize Adaptation Phase};

\node[process] (cont) at (8.6, -2.2) {\textbf{Contrastive}\\[1mm]\footnotesize Adaptation Phase};

\node[process] (seqcont) at (13.0, 0) {\textbf{Contrastive}\\[1mm]\footnotesize Sequential\\\footnotesize Adaptation Phase};

\node[final model, fill=teal!15] (bimntp) at (17.8, 2.2) {\textbf{Bi+MNTP (3)}\\[1mm]\footnotesize Adapted Weights\\\footnotesize Bidirectional Attention};

\node[final model, fill=teal!15] (bicomb) at (17.8, 0) {\textbf{Bi+MNTP+Contrastive (5)}\\[1mm]\footnotesize Sequential Adapted Weights\\\footnotesize Bidirectional Attention};

\node[final model, fill=teal!15] (bicont) at (17.8, -2.2) {\textbf{Bi+Contrastive (4)}\\[1mm]\footnotesize Adapted Weights\\\footnotesize Bidirectional Attention};


\draw[->] (base.east) -- (bibase.west);

\draw[->] (bibase.east) -- ++(0.4,0) |- (recon.west);
\draw[->] (bibase.east) -- ++(0.4,0) |- (cont.west);

\draw[->] (recon.east) -- (bimntp.west);
\draw[->] (recon.east) -- ++(0.4,0) |- (seqcont.west);

\node[right, font=\footnotesize\itshape, text=black!80] at (10.7, 1.1) {followed by};

\draw[->] (cont.east) -- (bicont.west);
\draw[->] (seqcont.east) -- (bicomb.west);

\end{tikzpicture}%
}
\caption{\textbf{Base (1):} The original causal model. \textbf{Bi+Base (2):} The Base model with bidirectional attention enabled. \textbf{Bi+MNTP (3):} The Bi+Base model with an MNTP adaptation phase. \textbf{Bi+Contrastive (4):} The Bi+Base model with a contrastive adaptation phase. \textbf{Bi+MNTP+Contrastive (5):} The Bi+Base model adapted sequentially using MNTP followed by contrastive training. Intermediate dashed blocks denote adaptation phases.}
\label{fig:model_taxonomy}
\end{figure}

\paragraph{Models and adaptation objectives.} We adapt two causal language model families initialized from pretrained weights: Gemma3 (270M, 1B)~\citep{gemmateam2025gemma3technicalreport} and Qwen3 (600M, 1.7B)~\citep{yang2025qwen3technicalreport}.\footnote{Models utilized: \href{https://huggingface.co/google/gemma-3-270m}{Gemma3-270M}, \href{https://huggingface.co/google/gemma-3-1b-pt}{Gemma3-1B}, \href{https://huggingface.co/Qwen/Qwen3-0.6B-Base}{Qwen3-0.6B}, and \href{https://huggingface.co/Qwen/Qwen3-1.7B-Base}{Qwen3-1.7B}; \autoref{sec:architecture_details}.} We use the smaller models for ablation studies and the larger models for scaling analysis, covering typical embedding model sizes. From these base models, we derive five distinct variants (detailed in \autoref{fig:model_taxonomy}) by switching from causal to bidirectional attention and applying two core adaptation objectives either individually or sequentially: Masked Next-Token Prediction (MNTP)~\citep{behnamghader2024llm2veclargelanguagemodels} and InfoNCE contrastive training~\citep{oord2019representationlearningcontrastivepredictive}.\footnote{Loss definitions and adaptation hyperparameters are provided in \autoref{sec:details_on_adaptation_objectives}.}

\paragraph{Adaptation corpus.} All adaptation experiments rely exclusively on open-source datasets. For clarity, we structure our corpora along three distinct domains:

\begin{enumerate}[leftmargin=15pt, topsep=0pt, label=\textbf{\arabic*.}]
    \item \textbf{English:} Masking uses FineWeb-Edu~\citep{penedo2024finewebdatasetsdecantingweb}; contrastive training uses the English subset of KaLM-embedding~\citep{zhao2025kalmembeddingv2superiortrainingtechniques} (7 hard negatives per query).
    
    \item \textbf{Multi-domain:} Masking relies on FineWeb-Edu (English), FineWeb2-HQ (multilingual, 20 languages)~\citep{messmer2026enhancingmultilingualllmpretraining}, FineMath~\citep{liu2024finemathfinegrainedmathematicalevaluation} (mathematics), and Stack V2 (code, 34 languages)~\citep{lozhkov2024starcoder2stackv2}. Contrastive training employs a merged corpus of 89 datasets (1 to 7 hard negatives per query) detailed in \autoref{sec:details_on_adaptation_data}.
    
    \item \textbf{Multimodal:} We introduce \textit{Omni-Contrastive},\footnote{Dataset available at: \url{https://huggingface.co/datasets/BidirLM/BidirLM-Omni-Contrastive}} a 1.8M-pair contrastive corpus mixing 65\% text-text (from the multi-domain corpus), 17.5\% audio-text from Laion-Audio-300M (200K, audio-description) and LibriSpeech ASR (100K, speech-transcription), and 17.5\% image-text from Colpali~\citep{faysse2024colpaliefficientdocumentretrieval} (100K, document-query), NatCap~\citep{teiletche2025modernvbertsmallervisualdocument}, and MSCOCO~\citep{lin2015microsoftcococommonobjects} (100K each, image-description).
\end{enumerate}

\paragraph{Evaluation protocol.} To reflect the current usage landscape, we assess encoder performance across diverse representation tasks under two distinct paradigms:

\begin{enumerate}[leftmargin=15pt, topsep=0pt, label=\textbf{\arabic*.}]
    \item \textbf{Fine-tuning evaluation:} We apply full-parameter adaptation for downstream tasks spanning the XTREME benchmark~\citep{hu2020xtrememassivelymultilingualmultitask} and four specific task categories: Information Retrieval (IR) via MIRACL~\citep{zhang_miracl_2023} and CodeSearchNet~\citep{husain2020codesearchnetchallengeevaluatingstate}; Sequence Classification (SC) via MNLI~\citep{mnli}, XNLI~\citep{conneau_xnli_2018}, PAWS-X~\citep{yang-etal-2019-paws}, MathShepherd~\citep{mathshepherd}, and CodeComplexity~\citep{jeon2023deep}; Token Classification (TC) via PAN-X and POS~\citep{hu2020xtrememassivelymultilingualmultitask}; and Sequence Regression (SR) via Seahorse~\citep{clark_seahorse_2023}.\footnote{Hyperparameters for fine-tuning and extensive dataset descriptions are provided in \autoref{apd_eval_data}.}
    
    \item \textbf{Embedding evaluation:} We assess off-the-shelf embedding performance via zero-shot and linear probing on MTEB-style benchmarks. Text evaluation uses English and Multilingual MTEB v2~\citep{muennighoff_mteb_2023,enevoldsen2025mmtebmassivemultilingualtext}, while cross-modal capabilities rely on MIEB lite~\citep{xiao2025miebmassiveimageembedding} (image-only, image-to-image, text-to-image) and MAEB beta~\citep{assadi2026maebmassiveaudioembedding} (audio-only, audio-to-audio, audio-text).
\end{enumerate}

\paragraph{Causal ecosystem.} To verify the capacity to compose our encoder with specialized variants from the causal ecosystem, we perform post-adaptation specialization across three domains:
\begin{enumerate}[leftmargin=15pt, topsep=0pt, label=\textbf{\arabic*.}]
    \item \textbf{Safety moderation:} We use Qwen3Guard-Gen-0.6B~\citep{zhao2025qwen3guardtechnicalreport} to transfer safety moderation knowledge to our encoder, assessed via safe/unsafe classification on the Beaver~\citep{ji2023beavertailsimprovedsafetyalignment}, Safe~\citep{ji2025pkusaferlhfmultilevelsafetyalignment}, and Aegis~\citep{ghosh2025aegis20diverseaisafety} datasets.
    
    \item \textbf{Vision:} Qwen3-VL-2B-Instruct~\citep{bai2025qwen3vltechnicalreport} transfers visual-textual knowledge, evaluated via visual-textual entailment on the e-SNLI-VE~\citep{do2021esnlivecorrectedvisualtextualentailment} benchmark.
    
    \item \textbf{Audio:} Qwen3-ASR-0.6B~\citep{shi2026qwen3asrtechnicalreport} transfers audio understanding, which we evaluate via textual comprehension with vocal questions on the BoolQ dataset.
\end{enumerate}

\section{Adaptation Strategies}\label{sec:Adaptation_Strategies}
\begin{figure}[h]
\centering
\includegraphics[width=\textwidth]{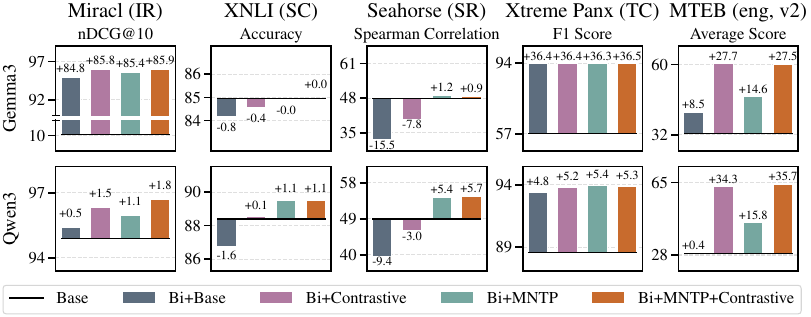}
\caption{\textbf{Performance comparison of model variants across downstream tasks.} Bars illustrate the absolute performance change relative to the unmodified Base model. Exact point differences are annotated above or below each bar.}
\label{fig:training_method_delta}
\end{figure}

Recent adaptation methods often rely exclusively on contrastive training, omitting masking objectives or bidirectional attention. To disentangle these choices,\footnote{Additional comparisons of MNTP and traditional masking objectives are provided in \autoref{sec:additional_results}.} we evaluate our five adaptation variants (\autoref{fig:model_taxonomy}) on Gemma3-270M and Qwen3-0.6B. We adapt these models using 10B tokens for masking and 3M samples for contrastive training on the English corpus (\autoref{sec:experimental_setup}), and report their downstream performance against the causal baseline (Base) in \autoref{fig:training_method_delta}.

\paragraph{Bidirectional attention drives performance.} 
As shown in \autoref{fig:training_method_delta}, enabling bidirectional attention at the fine-tuning stage only (Bi+Base) produces mixed results: it improves token classification and retrieval across both architectures but degrades performance on XNLI and Seahorse. However, introducing an MNTP adaptation phase unlocks the full benefit of bidirectional attention, boosting performance across all tasks with notable gains on XNLI and Seahorse (Gemma: +0.8 and +9.0; Qwen: +2.7 and +8.4, respectively).

\paragraph{Masking and contrastive objectives are complementary.}
Consistent with prior work \citep{gao2021simcse, li2023generaltextembeddingsmultistage, behnamghader2024llm2veclargelanguagemodels}, contrastive objectives drive generic embedding performance under zero-shot and linear probing evaluation (\autoref{fig:training_method_delta}), outperforming Bi+MNTP on MTEB by over 13 points across both architectures. However, our controlled comparison reveals that contrastive training alone sacrifices performance on tasks requiring full-parameter fine-tuning (e.g., XNLI, Seahorse) for embedding gains. To leverage the strengths of both paradigms, we employ a sequential adaptation strategy: MNTP followed by contrastive training (Bi+MNTP+Contrastive). This approach matches or surpasses the individual objectives across all tasks.

\keyfinding{Recent contrastive-only adaptations sacrifice fine-tuning quality for embedding gains; restoring a prior MNTP phase enables achieving peak performance across both regimes.}

\section{Scaling Adaptation Phases}\label{sec:scaling}
Following \autoref{sec:Adaptation_Strategies}, we scale the adaptation process while aiming to preserve the foundational knowledge of the base models. However, training on corpora diverging from the original pre-training distribution inherently risks alignment drift and catastrophic forgetting.

\subsection{Catastrophic Forgetting}

\begin{figure}[h]
\centering
   \includegraphics[width=\textwidth]{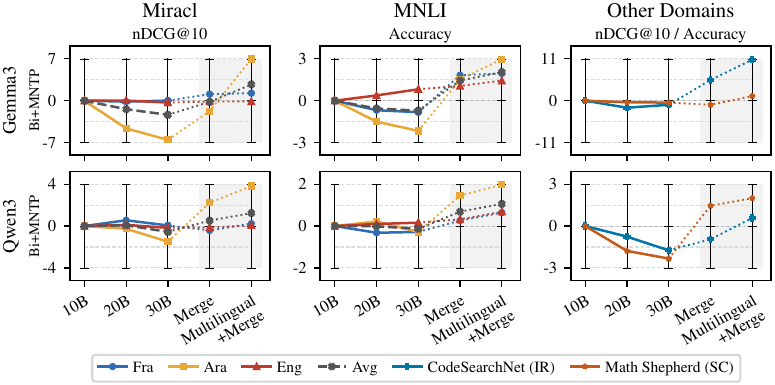}
\caption{\textbf{Evolution of model performances during long run adaptation.} Solid lines depict the absolute score change relative to the initial 10B adaptation, while dotted lines highlight the impact of complementary solutions to retain general knowledge.}
\label{fig:forgetting_interaction}
\end{figure}

To assess forgetting under realistic constraints (\autoref{fig:forgetting_interaction}), we extend MNTP adaptation for Gemma3-270M and Qwen3-0.6B from 10B to 30B tokens on the English corpus (\autoref{sec:experimental_setup}). Simultaneously, we monitor multi-domain performance at 10B-token intervals across multilingual (MIRACL, XNLI), code (CodeSearchNet), and math (Math Shepherd) benchmarks.

\textbf{Scaled Adaptation Induces Forgetting.}
As expected, scaling adaptation on a distribution unaligned with the original pre-training data leads to a clear forgetting phenomenon as training progresses: Gemma declines on Arabic (-7.0 points on MIRACL, -2.0 on XNLI), while Qwen demonstrates forgetting on Math Shepherd (-1.5) and CodeSearchNet (-2.0). To counteract this degradation, we propose two complementary approaches: a data-free model merging strategy and a lightweight multi-domain data mixture.

\textbf{Model Merging Mitigates Forgetting and Preserves Bidirectional Capabilities.} Motivated by the observation that the adapted and base models remain close in weight space, with an average cosine similarity of 0.78 for Gemma and 0.97 for Qwen,\footnote{We further analyze this finding in \autoref{sec:details_on_merging_and_similarities}, detailing the layer-wise similarity evolution.} we explore linear model merging~\citep{wortsman2022modelsoupsaveragingweights}, a technique shown to mitigate forgetting by averaging weights between different checkpoints. Specifically, we merge the 30B-token English-only MNTP models with their original base checkpoints using interpolation ratios ranging from 10\% to 90\% (30\% means a 0.3 weighting factor is applied to the base model).

\begin{figure}[h]
\centering
\includegraphics[width=\textwidth]{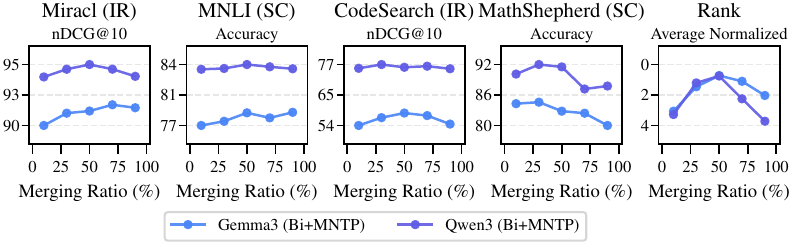}
\caption{\textbf{Model performance across merging ratios.} The first four columns report task scores, while the rightmost column reports the model ranking based on average normalized performance across all tasks. Merging ratio index dictates the interpolation weight.}
\label{fig:merging_ratio}
\end{figure}

As shown in \autoref{fig:merging_ratio}, performance peaks near a 50\% merging ratio, intuitively balancing the adapted bidirectional attention patterns with base model's distributional coverage. Consequently, we report the 30B adapted checkpoints at this 50\% ratio (denoted \textit{Merge}) in \autoref{fig:forgetting_interaction}. Compared to unmerged baselines, these models yield substantial cross-domain gains: +6 points on Arabic MNLI and code for Gemma, and +4 points in math for Qwen. Overall, merging emerges as a highly effective, data-free strategy to recover original knowledge.

\paragraph{Multi-domain data mixtures and weight merging yield optimal retention.} Complementing model merging, we investigate how multi-domain training mixtures mitigate adaptation forgetting without prior knowledge of the original pre-training distribution.

\begin{figure}[h]
\centering
\includegraphics[width=\textwidth]{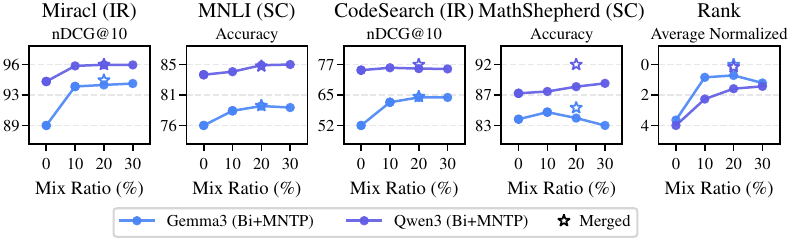}
\caption{\textbf{Model performance across data mix ratios.} The first four columns report task scores, while the rightmost column reports the model ranking based on average normalized performance across all tasks. The mix ratio specifies the proportion of multi-domain data.}
\label{fig:mix_multilingual}
\end{figure}

As illustrated in \autoref{fig:mix_multilingual}, we replace part of our initial English mix with multi-domain data (\autoref{sec:experimental_setup}) distributed equally across multilingual, math, and code domains. We observe that performance plateaus when allocating just 20\% to 30\% of the mixture to this multi-domain subset, indicating that a small fraction is sufficient to preserve original knowledge. To control for this factor, we fix this ratio at 20\% for subsequent experiments. Building on our merging strategy, interpolating this checkpoint with the original base weights yields further gains. This final \textit{Multilingual+Merge} configuration (\autoref{fig:forgetting_interaction}) achieves our best overall results, with an average improvement of +2 points on XNLI and MIRACL for both architectures, and up to +11 points on code benchmarks for Gemma.

\keyfinding{Combining weight merging with a lightweight multi-domain data mixture preserves the base model's foundational knowledge and newly acquired bidirectional capabilities.}

\section{Frontier Performance Through Scaled Adaptation}\label{sec:frontier_performance}
Building upon our best-performing adaptation strategies (\autoref{sec:Adaptation_Strategies}) and empirical findings (\autoref{sec:scaling}), we scale our approach to larger architectures, yielding four Bi+MNTP variants: Gemma3 (270M and 1B) and Qwen3 (0.6B and 1.7B). To establish strong general-purpose embedding capabilities, we execute the second step of our biphasic pipeline via contrastive training on 10M samples from our multi-domain corpus (\autoref{sec:experimental_setup}). We evaluate these final models, denoted the BidirLM series, on MTEB and an augmented XTREME benchmark (incorporating math and code domains, detailed in \autoref{apd_eval_data}), plotting the Pareto frontier against the latest fully open-source models (i.e., those releasing complete contrastive training data).

\begin{figure}[h]
\centering
\includegraphics[width=\textwidth]{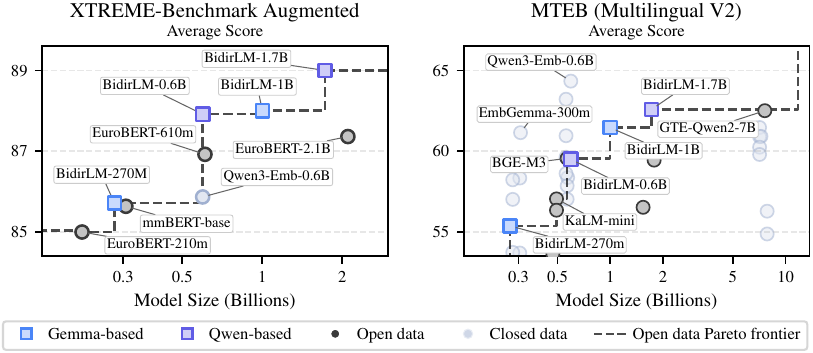}
\caption{\textbf{Multilingual model performance by size.} We report the average scores of the latest multilingual models across individual tasks on the XTREME and MTEB benchmarks. The dashed line indicates the open-source performance Pareto frontier.}
\label{fig:mteb_multi_v2}
\end{figure}

\paragraph{Adapted models redefine the Pareto frontier on task-specific benchmarks.} Under full-parameter fine-tuning, all BidirLM variants establish a new performance frontier on the augmented XTREME benchmark. Notably, BidirLM-270M matches the performance of mmBERT-base \citep{marone2025mmbertmodernmultilingualencoder} while utilizing 10\% fewer parameters, and BidirLM-0.6B outperforms its closest counterpart (EuroBERT-610m) by more than 1 point.\footnote{Models such as BGE-M3, KaLM, and EmbedGemma couldn't be evaluated due to their lack of architectural support for sentence or token classification, a key limitation of embedding-only models.}

\paragraph{Adapted models redefine the open-source Pareto frontier on generic embedding tasks.} Traditionally, generic embeddings and task-specific fine-tuning rely on separate model variants. Our adaptation eliminates this trade-off: beyond achieving the strongest performance on full-parameter fine-tuning, the exact same BidirLM variants advance the open-source Pareto frontier across three of our four size configurations on generic embedding benchmarks (MTEB). Notably, we accomplish this using only classical contrastive training, completely avoiding knowledge distillation from proprietary models or costly multi-run averaging. Consequently, our models constitute robust open-source baselines for future work challenging closed-source systems such as Qwen3-Embedding and EmbeddingGemma.

\section{Domain and Modality Specialization}\label{sec:alignment_and_domain_specificity}
Motivated by the observation that weight merging efficiently preserves the base model's foundational knowledge and bidirectional capabilities (\autoref{sec:scaling}), we push the boundaries of this technique to tailor our generic encoders to new domains and modalities, harnessing the vast ecosystem of specialized generative models.

\subsection{Domain Alignment}
We explore domain knowledge transfer by exploiting the shared backbone between our Bi+MNTP Qwen3-0.6B and the Qwen3Guard-Gen-0.6B safety model (\autoref{sec:experimental_setup}). We merge them at a 50\% ratio\footnote{We provide a detailed analysis for merge ratios $\in \{0, 0.25, 0.5, 0.75, 1\}$ in \autoref{sec:merging_ratio_causal_specialists}} (cos sim: 0.97) and perform 500 fine-tuning steps on the Beaver training set (two minutes on one MI250X GPU). We evaluate the resulting encoder on the Beaver test set and two out-of-distribution benchmarks (Safe and Aegis) against the specialist causal model fine-tuned with bidirectional attention (Bi+Specialist) and the Bi+MNTP models.

\begin{figure}[h]
\centering
\includegraphics[width=\textwidth]{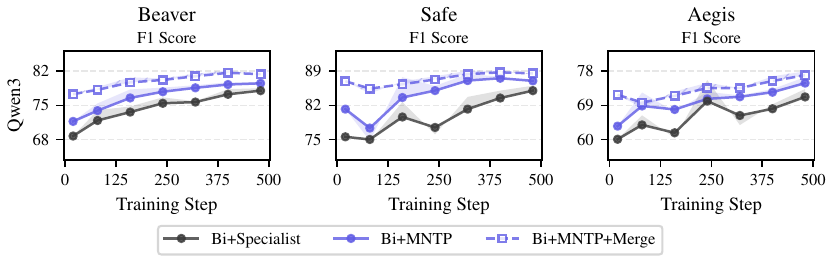}
\caption{\textbf{Evolution of performance during domain specialization}. We report test split performance on Beaver, SAFE and Aegis. Solid lines correspond to the exponential moving averaged (EMA) curves ($\alpha = 0.4$), with shaded areas showing raw value deviation.}
\label{fig:guard_benchmarks_curves}
\end{figure}

\paragraph{Merged model outperforms all baseline configurations.} As shown in \autoref{fig:guard_benchmarks_curves}, the merged encoder (Bi+MNTP+Merge) outperforms all other configurations by over 1 point on average. It also shows better out-of-distribution generalization and greater training stability, with minimal variance between raw measurements and EMA-smoothed curves.

\paragraph{Merging enables rapid sample-efficient adaptation.} The Bi+MNTP+Merge model reaches over 93\% of its peak performance across all benchmarks in just 20 steps (80 samples). At this early training stage, it outperforms all other variants by a margin of more than 5 points.

\subsection{Modality Alignment}
We extend this approach to new modalities by merging the bimodal vision-text Qwen3-VL-2B-Instruct and the unimodal audio Qwen3-ASR-0.6B models with our adapted Bi+MNTP encoders (Qwen3-1.7B and Qwen3-0.6B, respectively) in equal proportions (cosine similarities: 0.97 for vision, 0.93 for audio). Finally, we conduct a 500-step fine-tuning phase on e-SNLI-VE (visual-textual entailment) and BoolQ-Audio (vocal comprehension) (\autoref{sec:experimental_setup}).

\begin{figure}[h]
\centering
\includegraphics[width=\textwidth]{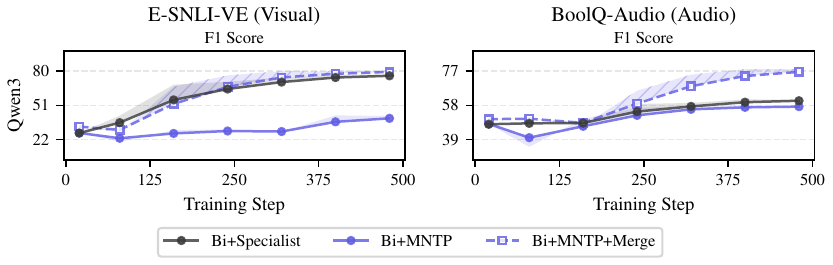}
\caption{\textbf{Evolution of performance during modality specialization}. We report test split F1 score on e-SNLI-VE (vision) and Boolq-Audio (Audio). Solid lines correspond to exponential moving average curves ($\alpha = 0.4$), with shaded areas showing raw data deviation.}
\label{fig:multimodal}
\end{figure}

\textbf{Modality Adaptation Reveals a Warm-Up Phase.} Merged variants yield the highest overall performance (\autoref{fig:multimodal}), exceeding Bi+Specialist by over 1 and 15 points on vision and audio tasks, and surpassing unmerged baselines by over 30 and 19 points respectively. Unlike in domain adaptation, the merged variant exhibits an initial warm-up period of 100 vision steps and 175 audio steps. Consistent with prior literature, this warm-up phase stems from the requirement to align internal representations with the newly introduced modality heads.

\textbf{Merging succeeds without shared modalities.} We observe a clear gap between baseline performances. While the Bi+Specialist remains competitive in vision, trailing the merged model by only 1 point, it degrades significantly in audio. We attribute this to the input modalities of the specialist models: the vision model already possessed multimodal capabilities for text and vision, whereas the audio model was trained exclusively for unimodal speech recognition. Crucially, we observe that merging technique still succeeds, demonstrating that models can be effectively combined even when they share no prior overlapping modalities.

\subsection{Omnimodal Alignment}
\begin{figure}[h]
\centering
\includegraphics[width=\textwidth]{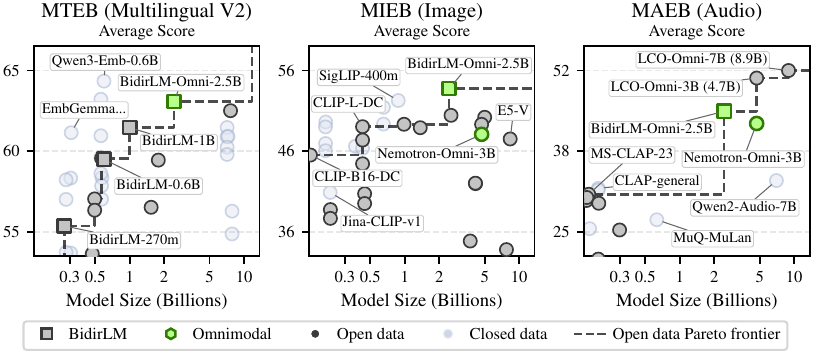}
\caption{\textbf{Embedding model performance by size.} Average score across individual tasks on MTEB Multilingual V2, MIEB, and MAEB, as a function of model size. The dashed line shows the Pareto frontier over open training data models.}
\label{fig:results_omni}
\end{figure}

Building on the observation that our encoder can easily adapt to new modalities, we introduce BidirLM-Omni-2.5B, a compact omnimodal model. We construct this by merging the textual backbones of three Qwen3-1.7B variants (ASR, VL, and Bi+MNTP) in equal proportions, appending their respective audio and visual heads (\autoref{sec:omni_diagram}). Following contrastive training on our multimodal corpus (\autoref{sec:experimental_setup}), we evaluate the model against numerous baselines (\autoref{fig:results_omni}) across MTEB (Text), MIEB (Image), and MAEB (Audio).

\paragraph{BidirLM-Omni sets a new omnimodal state of the art.} BidirLM-Omni-2.5B outperforms the latest best-performing omnimodal model, Nemotron-Omni-3B~\citep{xu2025omniembednemotronunifiedmultimodalretrieval}, across all modalities, achieving notable gains on text (+17) and image (+5) benchmarks while being nearly half the size (2.5B vs. 4.8B).

\paragraph{BidirLM-Omni surpasses unimodal specialists several times larger.} Beyond outperforming its omnimodal counterparts, BidirLM-Omni-2.5B establishes new Pareto frontiers regardless of data transparency. Notably, the merging process efficiently leverages the strengths of each model variant, enabling it to rank first among all baselines on the MIEB benchmark and third on MAEB, surpassing bimodal architectures many times its size.

\paragraph{Composing specialized models yields efficient and flexible encoders.} By reusing existing specialized models rather than training from scratch, BidirLM-Omni required only 250 additional GPU hours (MI250X) of compute for merging and contrastive training, demonstrating that new omnimodal architectures can be assembled incrementally as specialized models become available, bypassing the need to retrain the entire pipeline.

\keyfinding{Weight merging enables the efficient composition of domain and modality causal model specialists into an adapted encoder, succeeding even in the absence of shared prior modalities.}

\section{Related Work}
\paragraph{Adapting Causal Models for Generic and Multimodal Representations.} Causal LLMs have emerged as strong backbones for text embeddings~\citep{ma2023finetuningllamamultistagetext,li-etal-2024-llama2vec,springer2025repetitionimproveslanguagemodel}, with adaptation strategies generally falling into two paradigms. The first injects bidirectionality through masking-based objectives, using either classical masked language modeling (MLM)~\citep{devlin2019bertpretrainingdeepbidirectional} or the next-token variant MNTP~\citep{behnamghader2024llm2veclargelanguagemodels}, prior to fine-tuning. While \citet{behnamghader2024llm2veclargelanguagemodels} first proposed the MNTP-then-contrastive pipeline, their evaluation did not isolate the contributions of bidirectional attention, the masking objective, and contrastive training itself. Consequently, the second paradigm, which is now dominant in practice, skips the masking phase entirely and applies contrastive learning directly~\citep{Le_Khac_2020, wang2024improvingtextembeddingslarge, lee2025nvembedimprovedtechniquestraining}. Within this approach, attention design varies: some methods enable full bidirectionality (e.g., Embedding-Gemma~\citep{vera2025embeddinggemmapowerfullightweighttext}), while others preserve causal masking (e.g., Qwen3 Embedding~\citep{zhang2025qwen3embeddingadvancingtext}). This contrastive paradigm has recently been extended to multimodal representations, yielding models like VLM2Vec~\citep{jiang2025vlm2vectrainingvisionlanguagemodels} and Nemotron-Omni~\citep{xu2025omniembednemotronunifiedmultimodalretrieval}.

\paragraph{Weight Merging for Knowledge Transfer and Specialization.} Adapting models to new objectives and distributions inevitably risks catastrophic forgetting \citep{french1999catastrophic}. While traditional continual learning relies on compute-intensive replay buffers or regularization \citep{rolnick2019experiencereplaycontinuallearning, wang2024comprehensivesurveycontinuallearning}, post-hoc weight merging (Model Soups~\citep{wortsman2022modelsoupsaveragingweights} or Task Arithmetic~\citep{ilharco2023editingmodelstaskarithmetic}) offers a highly efficient alternative, enabling models to seamlessly adapt to new distributions. However, these techniques have historically been applied to models sharing similar objectives and attention mechanisms.

\section{Conclusion}In this work, we introduce a unified, fully open-source framework for transforming causal decoder LLMs into bidirectional encoders spanning text to multiple modality domains. Through systematic comparisons, we show that the masking phase omitted by recent contrastive-only methods is in fact critical for fine-tunig performance. To scale this adaptation without proprietary pre-training data, we employ a dual strategy of linear weight merging and a lightweight multi-domain data mixture, yielding the BidirLM model family. Rather than building inflexible systems, our framework seamlessly composes specialized generative models with our adapted encoders, enabling efficient cross-modal and domain-specific adaptation without retraining entire pipelines, culminating in BidirLM-Omni.

\section*{Future Work}
\paragraph{Contrastive training.} Our ablations focused on the masking phase, a step frequently omitted in concurrent work. While contrastive training already benefits from an extensive body of prior work and ablations~\citep{xu2025omniembednemotronunifiedmultimodalretrieval, zhang2025qwen3embeddingadvancingtext, vera2025embeddinggemmapowerfullightweighttext, hu2025kalmembeddingsuperiortrainingdata}, systematically studying data composition, hard-negative mining strategies, and scaling behavior in the omnimodal setting remains a natural next step.

\paragraph{Additional mitigation techniques and model architectures.} In this study, we rely on linear merging and data mixing, both lightweight by design. We plan to explore richer regularization strategies, notably knowledge distillation~\citep{hinton2015distillingknowledgeneuralnetwork} from the base model. Utilizing recent techniques that enable cross-architectural distillation~\citep{boizard2025crosstokenizerdistillationuniversallogit, minixhofer2025universalcrosstokenizerdistillationapproximate} may offer stronger knowledge retention at the cost of additional compute. Finally, validating our framework on non-transformer causal architectures, such as state-space models~\citep{gu2024mambalineartimesequencemodeling, yang2025gateddeltanetworksimproving}, remains an open question.

\section*{Acknowledgments}
We sincerely thank the ADASTRA supercomputer (CINES) for its high-performance computing (HPC) resources, provided through grant A0181016236. This work was also supported by the Jean Zay supercomputer (GENCI-IDRIS-CNRS) through grant AD010617149, and the ROMEO HPC center at the University of Reims. Furthermore, we gratefully acknowledge the support of the French government through the France 2030 program as part of the ArGiMi project.

\clearpage

\bibliography{colm2026_conference}
\bibliographystyle{colm2026_conference}

\appendix
\clearpage

\section{Base Model Architecture Details}\label{sec:architecture_details}

This appendix summarizes the two causal language model families used throughout this work in \autoref{tab:arch_comparison}: Gemma3~\citep{gemmateam2025gemma3technicalreport} and Qwen3~\citep{yang2025qwen3technicalreport}. Both families follow a decoder-only transformer design but differ in architectural choices such as attention patterns, normalization layers, vocabulary sizes, and pre-training configurations, providing evidence that our framework generalizes across diverse causal decoder architectures.

\begin{table*}[h]
\centering
\caption{Architectural comparison of base models used in this work.}
\label{tab:arch_comparison}
\small
\begin{tabular}{@{}lcccc@{}}
\toprule
& \textbf{Gemma3-270M} & \textbf{Gemma3-1B} & \textbf{Qwen3-0.6B} & \textbf{Qwen3-1.7B} \\
\midrule
\multicolumn{5}{@{}l}{\textit{Architecture}} \\
Parameters           & 268M  & 1001M & 596M   & 1721M \\
Layers               & 18    & 26    & 28    & 28    \\
Hidden dimension     & 640   & 1152  & 1024  & 2048  \\
Normalization        & \multicolumn{2}{c}{RMSNorm (pre \& post attention)}     & \multicolumn{2}{c}{RMSNorm (pre attention)} \\
Embedding tying      & \multicolumn{2}{c}{No}                        & \multicolumn{2}{c}{Yes} \\
Attention pattern    & \multicolumn{2}{c}{5:1 sliding / global}      & \multicolumn{2}{c}{Uniform global} \\
Head dimension       & \multicolumn{2}{c}{256}                       & \multicolumn{2}{c}{128} \\
Attention heads      & 4     & 4     & 16    & 16    \\
KV heads             & 1     & 1     & 8     & 8     \\
Sliding-window span  & \multicolumn{2}{c}{512 tokens}                & \multicolumn{2}{c}{---} \\
RoPE base $\theta$   & \multicolumn{2}{c}{10k (local) / 1M (global)} & \multicolumn{2}{c}{1M} \\
\midrule
\multicolumn{5}{@{}l}{\textit{Pre-training}} \\
Tokenizer            & \multicolumn{2}{c}{SentencePiece}             & \multicolumn{2}{c}{Byte-level BPE} \\
Vocab size           & \multicolumn{2}{c}{262,144}                   & \multicolumn{2}{c}{151,936} \\
Pre-training tokens  & 6~T   & 2~T   & \multicolumn{2}{c}{36~T} \\
Distillation         & \multicolumn{2}{c}{Yes}                       & \multicolumn{2}{c}{No} \\
\bottomrule
\end{tabular}
\end{table*}

\section{Adaptation Training Details}\label{sec:details_on_adaptation_objectives}

\subsection{Loss definitions}
\begin{enumerate}[leftmargin=15pt]
    \item \textbf{Masked Language Modeling (MLM).} A subset of tokens is randomly masked, and the model is trained to reconstruct them using full bidirectional context:
    \begin{equation}
        \mathcal{L}_{\text{MLM}}(\mathbf{x}) = -\sum_{i \in \mathcal{M}} \log p_{\theta}\!\left(x_i \mid \mathbf{x}_{\mathcal{M}}\right),
    \end{equation}
    where $\mathcal{M} \subset \{1,\dots,T\}$ denotes the masked positions and $\mathbf{x}_{\mathcal{M}}$ is the input sequence with masked tokens replaced by a special \texttt{[MASK]} placeholder. Masking is applied independently with probability $p_{\text{mask}} \in \{10\%, 20\%, 30\%, 40\%\}$, which we evaluate in \autoref{sec:additional_results}.

    \item \textbf{Masked Next-Token Prediction (MNTP).} MNTP combines masked reconstruction with the causal next-token prediction mechanism by predicting each masked token $x_i$ from the logits at position $i-1$:
    \begin{equation}
        \mathcal{L}_{\text{MNTP}}(\mathbf{x}) = -\sum_{i \in \mathcal{M}} \log p_{\theta,\, i-1}\!\left(x_i \mid \mathbf{x}_{\mathcal{M}}\right).
    \end{equation}
    All masking-related notation and hyperparameters follow the MLM setup.

    \item \textbf{Contrastive learning (InfoNCE).} We employ a contrastive objective with both in-batch and hard negatives to align the representations of semantically equivalent sequences. For each anchor $\mathbf{x}$ and positive $\mathbf{x}^+$, the negatives $\mathcal{N}$ consist of the remaining in-batch samples, augmented with explicitly mined hard negatives:
    \begin{equation}
        \mathcal{L}_{\text{InfoNCE}} = -\log \frac{e^{\operatorname{sim}(\mathbf{h}_{\mathbf{x}}, \mathbf{h}_{\mathbf{x}^+}) / \tau}}{e^{\operatorname{sim}(\mathbf{h}_{\mathbf{x}}, \mathbf{h}_{\mathbf{x}^+}) / \tau} + \sum_{\mathbf{x}^- \in \mathcal{N}} e^{\operatorname{sim}(\mathbf{h}_{\mathbf{x}}, \mathbf{h}_{\mathbf{x}^-}) / \tau}},
    \end{equation}
    where $\mathbf{h}_{\mathbf{x}} = f_\theta(\mathbf{x})$ denotes the sequence representation, obtained either via last-token selection or mean pooling over the final layer hidden states, $\operatorname{sim}(\cdot,\cdot)$ is the cosine similarity and $\tau$ is a temperature hyperparameter.
\end{enumerate}

\subsection{Hyperparameters}
To ensure a strictly controlled and fair comparison, all training runs process identical data in the exact same order for a single epoch of unique tokens during masking, and unique sentence pairs during contrastive adaptation. Learning rates (LR) are chosen via grid search over 10 log-spaced values from $1\times10^{-5}$ to $1\times10^{-3}$, selecting the value that minimizes training loss on 1B tokens for masking and 1M samples for contrastive training. All experiments use a fixed seed (42) for reproducibility.

\begin{table}[h]
\centering
\caption{Masked adaptation hyperparameters.}
\label{tab:hyperparams_masked}
\small
\begin{tabular}{@{}lcccc@{}}
\toprule
& \textbf{Gemma3-270M} & \textbf{Qwen3-0.6B} & \textbf{Gemma3-1B} & \textbf{Qwen3-1.7B} \\
\midrule
Learning rate           & $5 \times 10^{-4}$ & $1 \times 10^{-4}$ & $7 \times 10^{-5}$ & $5 \times 10^{-5}$ \\
Warmup steps            & \multicolumn{4}{c}{0.01 epoch}  \\
LR scheduler            & \multicolumn{4}{c}{Warmup Stable Decay (WSD)~\citep{hu2024minicpmunveilingpotentialsmall}} \\
Optimizer               & \multicolumn{4}{c}{AdamW (fused)} \\
Max grad norm           & \multicolumn{4}{c}{1.0} \\
Seed                    & \multicolumn{4}{c}{42} \\
\midrule
Batch size (per GPU)    & \multicolumn{4}{c}{1} \\
Gradient accumulation   & \multicolumn{4}{c}{3} \\
Num GPUs                & \multicolumn{4}{c}{96 (12 nodes $\times$ 8 GPUs)} \\
Sequence length         & \multicolumn{4}{c}{8192} \\
Effective batch size    & \multicolumn{4}{c}{2,359,296} \\
\midrule
Adaptation objective    & \multicolumn{4}{c}{Masked Next Token Prediction (MNTP)} \\
Loss function           & \multicolumn{4}{c}{Linear Cross Entropy (fused)} \\
\bottomrule
\end{tabular}
\end{table}

\begin{table}[h]
\centering
\caption{Contrastive training hyperparameters.}
\label{tab:hyperparams_contrastive}
\small
\begin{tabular}{@{}lcccc@{}}
\toprule
& \textbf{Gemma3-270M} & \textbf{Qwen3-0.6B} & \textbf{Gemma3-1B} & \textbf{Qwen3-1.7B} \\
\midrule
Learning rate           & $6 \times 10^{-5}$ & $3 \times 10^{-5}$ & $1 \times 10^{-4}$ & $1 \times 10^{-4}$ \\
Warmup steps            & \multicolumn{4}{c}{500} \\
LR scheduler            & \multicolumn{4}{c}{Linear} \\
Optimizer               & \multicolumn{4}{c}{AdamW (fused)} \\
Max grad norm           & \multicolumn{4}{c}{1.0} \\
Seed                    & \multicolumn{4}{c}{42} \\
\midrule
Batch size (per GPU)    & \multicolumn{4}{c}{128} \\
Num GPUs                & \multicolumn{4}{c}{4} \\
Effective batch size    & \multicolumn{4}{c}{512} \\
Mini-batch size         & 64    & 32    & 32    & 16    \\
\midrule
Loss function           & \multicolumn{4}{c}{Cached Multiple Negatives Ranking Loss} \\
Temperature $\tau$      & \multicolumn{4}{c}{0.05} \\
In-batch negatives      & \multicolumn{4}{c}{Yes} \\
Hard negatives          & \multicolumn{4}{c}{1--7 (0 for multimodal samples)} \\
\midrule
Max sequence length     & \multicolumn{4}{c}{512} \\
\bottomrule
\end{tabular}
\end{table}

\section{Adaptation Data Details}\label{sec:details_on_adaptation_data}
\paragraph{Masking:}
\begin{itemize}[leftmargin=15pt, itemsep=1pt, topsep=0pt]
    \item \textbf{FineWeb-Edu~\citep{penedo2024finewebdatasetsdecantingweb}} consists of 1.3T English tokens from educational web pages filtered from the FineWeb dataset.\footnote{\url{https://huggingface.co/datasets/HuggingFaceFW/fineweb-edu}}

    \item \textbf{FineWeb2-HQ~\citep{messmer2026enhancingmultilingualllmpretraining}} is a high-quality, model-filtered pretraining dataset derived as a subset of FineWeb2, spanning 20 languages. It was created by selecting the top 10\% of documents in each language based on scores from a deep learning classifier trained to identify structured, knowledge-rich samples.\footnote{\url{https://huggingface.co/datasets/epfml/FineWeb2-HQ}}
    
    \item \textbf{FineMath~\citep{liu2024finemathfinegrainedmathematicalevaluation}} comprises 54B tokens of mathematical content filtered from CommonCrawl to retain only the most educational material, focusing on clear explanations and step-by-step problem-solving.\footnote{\url{https://huggingface.co/datasets/HuggingFaceTB/finemath}}
    
    \item \textbf{The Stack v2~\citep{lozhkov2024starcoder2stackv2}} contains over 3B files across 600+ programming and markup languages.\footnote{\url{https://huggingface.co/datasets/bigcode/the-stack-v2}}
\end{itemize}

\paragraph{Contrastive:}

\begin{table*}[h]
\centering
\sisetup{group-separator={,}, group-minimum-digits=4}
\caption{Training dataset composition after domain decontamination (\num{10110219} training pairs).}
\label{tab:dataset_breakdown}

\small
\setlength{\tabcolsep}{4pt}

\begin{tabular}{@{}l S[table-format=7.0]@{\hspace{14pt}}l S[table-format=7.0]@{}}
\toprule
\textbf{Dataset} & \textbf{Pairs} & \textbf{Dataset} & \textbf{Pairs} \\
\midrule

\multicolumn{2}{@{}l}{\textsc{KaLM}} & \multicolumn{2}{l}{} \\[2pt]
mMARCO (zh)             & 379870 & NLLB                    & 26504 \\
SimCLUE                 & 290699 & ESCI                    & 26043 \\
Multi-CPR               & 234587 & Aya Dataset             & 22449 \\
SimCSE NLI              & 217099 & Yahoo Answers           & 21724 \\
T2Ranking               & 188606 & CSL                     & 19945 \\
nli\_zh                 & 185787 & LCSTS                   & 19535 \\
llm\_retr.\_short\_long & 149511 & THUCNews                & 19288 \\
llm\_sts\_monolingual   & 132561 & WebGPT Comparisons      & 18924 \\
CMNLI                   & 119029 & ChatMed-Dataset         & 18608 \\
llm\_retr.\_long\_long  & 114979 & AdvertiseGen            & 17526 \\
llm\_retr.\_long\_short & 114584 & OCNLI                   & 11937 \\
DuReader\_checklist     & 97764  & ATEC                    & 11387 \\
cMedQA-V2.0             & 88109  & BQ                      & 10000 \\
PubMedQA                & 79954  & SearchQA                & 9988 \\
DuReader                & 79229  & CMRC 2018               & 9753 \\
ELI5                    & 76408  & rag-dataset-12000       & 9272 \\
llm\_retr.\_short\_short & 76315 & lawzhidao               & 6784 \\
llm\_sts\_bitext\_retr. & 75271  & webqa                   & 4988 \\
XNLI (zh)               & 74252  & CHEF                    & 4824 \\
MEDI2BGE                & 71790  & cCOVID-News             & 4727 \\
MultiNLI                & 63701  & DRCD                    & 4714 \\
Natural Questions       & 56377  & AFQMC                   & 3876 \\
RefGPT                  & 49896  & CINLID                  & 2883 \\
CodeFeedback            & 49090  & UMETRIP-QA              & 2537 \\
WikiAnswers             & 47686  & ChineseSTS              & 2497 \\
QBQTC                   & 47223  & LIMA                    & 1991 \\
Mr.~TyDi                & 46997  & WebCPM                  & 1602 \\
OpenOrca                & 38623  & ExpertQA                & 1252 \\
retrieval\_data\_llm    & 32551  & CAIL2019-SCM            & 648 \\
MLDR                    & 31097  & ContractNLI             & 628 \\
CC-News                 & 28246  & law-gpt                 & 500 \\
\cmidrule(r){1-2} \cmidrule(l){3-4}
\multicolumn{2}{@{}l}{\textit{KaLM subtotal}} & & \textit{3655225} \\[4pt]
\multicolumn{2}{@{}l}{\textsc{Nemotron}} & \multicolumn{2}{l}{\textsc{Other}} \\[2pt]

SyntheticClassif.       & 1044212 & Parallel Data (51 lang.\ pairs) & 3054406 \\
PAQ                     & 1000000 & \quad OPUS-100          & 946599 \\
MS~MARCO                & 532751  & \quad JW300             & 701201 \\
MAmmoTH2                & 317180  & \quad TED Talks         & 733318 \\
NaturalQuestions         & 100231  & \quad WikiMatrix        & 673288 \\
GooAQ                   & 100000  & InF-IR           & 48403 \\
SQuAD                   & 87599  & \quad MS MARCO   & 38759 \\
MIRACL                  & 79648  & \quad metamath   & 7104 \\
TriviaQA                & 73346  & \quad leetcode   & 2540  \\
EmotionClassif.         & 13039  & FollowIR\          & 494  \\
NFCorpus                & 3685   &  &   \\
\cmidrule(r){1-2} \cmidrule(l){3-4}
\textit{Nemotron subtotal} & \textit{3351691} & \textit{Other subtotal} & \textit{3103303} \\

\midrule
\multicolumn{3}{@{}l}{\textbf{Total}} & \textbf{10110219} \\
\bottomrule
\end{tabular}

\end{table*}

\begin{itemize}[leftmargin=15pt, itemsep=1pt, topsep=0pt]
    \item \textbf{Embed-Nemotron-Dataset-V1~\citep{babakhin2025llamaembednemotron8buniversaltextembedding}} is a curated subset of 11 datasets used for training general-purpose text embedding models.\footnote{\url{https://huggingface.co/datasets/nvidia/Embed-Nemotron-Dataset-V1}}
    \item \textbf{KaLM-embedding-finetuning-data \citep{hu2025kalmembeddingsuperiortrainingdata}} is a diverse collection of 62 datasets spanning retrieval, semantic textual similarity, and classification tasks.\footnote{\url{https://huggingface.co/datasets/HIT-TMG/KaLM-embedding-finetuning-data}}
    \item \textbf{Parallel Data} spans 51 English-centric language pairs (\num{3054406} pairs after subsampling; see Table~\ref{tab:dataset_breakdown}), drawn from 4 datasets; OPUS-100\footnote{\url{https://huggingface.co/datasets/Helsinki-NLP/opus-100}}, JW300\footnote{\url{https://huggingface.co/datasets/sentence-transformers/parallel-sentences-jw300}}, TED Talks\footnote{\url{https://huggingface.co/datasets/sentence-transformers/parallel-sentences-talks}}, and WikiMatrix\footnote{\url{https://huggingface.co/datasets/sentence-transformers/parallel-sentences-wikimatrix}}.
    \item \textbf{FollowIR~\citep{weller2024followirevaluatingteachinginformation}} is an instruction-aware retrieval training set, used here excluding Core17/News21/Robust04 evaluation topics.\footnote{\url{https://huggingface.co/datasets/jhu-clsp/FollowIR-train}}
    \item \textbf{InF-IR~\citep{zhuang2025betterinstructionfollowingretrieval}} provides instruction-following information retrieval training data, used here excluding Robust04 evaluation topics.\footnote{\url{https://huggingface.co/datasets/InF-IR/InF-IR}}
\end{itemize}

\paragraph{Instruction-aware training with single-domain batching.} For asymmetric tasks (retrieval, reranking), instruction prefixes are prepended to queries only; for symmetric tasks (STS, pair classification), the same instruction is applied to both anchors and positives. Each training batch is drawn from a single dataset so that all in-batch negatives share the query's exact task structure and domain. Furthermore, each query is paired with mined hard negatives ranging from 1 to 7 ensuring every sample retains at least one hard negative.

\paragraph{MTEB decontamination.} 
To ensure fair zero-shot evaluation on the MTEB benchmark~\citep{muennighoff_mteb_2023}, we exclude training domains that overlap with MTEB evaluation tasks. This removes 13~domain families (including both KaLM and Nemotron versions): ArXiv QA, MASSIVE (classification and clustering), CQADupstack, TREC-COVID, DBPedia, FEVER, FiQA, HotpotQA, PAWS-X, Quora, SciFact, and SNLI. The fully decontaminated corpus totals \num{10110219} training pairs (\autoref{tab:dataset_breakdown}).

\paragraph{Dataset deduplication.}
When merging the NeMo and KaLM sources, we adopt a \emph{NeMo-first} deduplication policy. Specifically, KaLM datasets that overlap with NeMo Retriever families (e.g., MIRACL, MS~MARCO, TriviaQA, SQuAD, NFCorpus, GooAQ, and PAQ) are dropped in favor of their Nemotron counterparts, which provide higher-quality hard negatives.

\section{Details of Evaluation}\label{apd_eval_data}

This appendix offers additional details on the datasets used for evaluation, they are organized into two sections: downstream task evaluation, where the model is fine-tuned on task-specific data, and zero-shot evaluation, where the model's frozen embeddings are evaluated directly (with at most a lightweight linear probe).

\subsection{Downstream Task Evaluation}

\paragraph{Sequence classification datasets (F1 score):}

\begin{itemize}[leftmargin=15pt, itemsep=1pt, topsep=0pt]
    \item \textbf{XNLI~\citep{conneau_xnli_2018} -- General:} This natural language inference task extends MNLI~\citep{mnli} to non-English languages, involving the classification of sentence pairs into entailment, contradiction, or neutral.\footnote{\url{https://huggingface.co/datasets/mteb/xnli}}

    \item \textbf{PAWS-X~\citep{yang-etal-2019-paws} -- General:} This dataset contains 23,659 human-translated Paraphrase Adversaries from Word Scrambling (PAWS) evaluation pairs across six distinct languages: French, Spanish, German, Chinese, Japanese, and Korean. The task aims to determine whether two sentences convey the exact same meaning.

    \item \textbf{MathShepherd~\citep{mathshepherd} -- Math:} This is a binary classification task aimed at determining whether a step-by-step math rationale is correct given a problem prompt.

    \item \textbf{CodeComplexity~\citep{jeon2023deep} -- Code:} This computational analysis task involves estimating the order of complexity for a code-formulated computer science problem.
\end{itemize}

\paragraph{Retrieval datasets (NDCG@10):}

\begin{itemize}[leftmargin=15pt, itemsep=1pt, topsep=0pt]
    \item \textbf{MS MARCO~\citep{bajaj_ms_2016} -- General:} This English-only retrieval dataset is used for fine-tuning. Each anchor--positive pair is augmented with a mined hard negative to form a triplet structure. We use the hard-triplet version of MS MARCO.\footnote{\url{https://huggingface.co/datasets/bclavie/msmarco-10m-triplets}}

    \item \textbf{MIRACL~\citep{zhang_miracl_2023} -- General:} For this multilingual retrieval dataset, we use the semi-supervised SentenceTransformers version as the primary data source.\footnote{\url{https://huggingface.co/datasets/sentence-transformers/miracl}} Anchors serve as queries, and the corpus consists of all positive documents in the dataset. Since only a single data split is available, we create validation and test sets by partitioning 50\% of the original split for each, using queries as the split key to ensure no data leakage.

    \item \textbf{CodeSearchNet~\citep{husain_codesearchnet_2019} -- Code:} This code retrieval dataset features comment--code query--positive pairs (SentenceTransformers version) and is processed similarly to the previous datasets.\footnote{\url{https://huggingface.co/datasets/sentence-transformers/codesearchnet}}
\end{itemize}

\paragraph{Sequence regression datasets (Spearman correlation):}

\begin{itemize}[leftmargin=15pt, itemsep=1pt, topsep=0pt]
    \item \textbf{SeaHorse~\citep{clark_seahorse_2023} -- Summary:} This multilingual summarization evaluation task annotates each text--summary pair across six binary dimensions. The final score is obtained by averaging these labels, yielding a continuous value between 0 and 1. To avoid penalizing models with limited context lengths, the summary is placed first in the input, followed by the main text, ensuring the model can attend to the full summary.\footnote{\url{https://huggingface.co/datasets/hgissbkh/seahorse}}
\end{itemize}

\paragraph{Token classification datasets (F1 score):}

\begin{itemize}[leftmargin=15pt, itemsep=1pt, topsep=0pt]
    \item \textbf{XTREME PAN-X~\citep{hu2020xtrememassivelymultilingualmultitask} -- NER:} Named entity recognition task which is a balanced subset of the \texttt{WikiAnn} dataset~\citep{pan-etal-2017-cross}. Named entities in Wikipedia were automatically annotated with \texttt{LOC}, \texttt{PER}, and \texttt{ORG} tags in IOB2 format using a combination of knowledge base properties, cross-lingual and anchor links, self-training, and data selection.\footnote{\url{https://huggingface.co/datasets/google/xtreme}}

    \item \textbf{XTREME POS~\citep{nivre2020universaldependenciesv2evergrowing} -- POS:} Cross-lingual structured prediction task requires assigning a grammatical category (noun, verb, adjective, etc.) to each token in a sentence. It uses data from Universal Dependencies v2.5, and models are evaluated under a zero-shot transfer setting: fine-tuned on English labeled data and directly applied to other languages without retraining. 
\end{itemize}

\paragraph{XTREME Augmented benchmark (Average score):} 
We create an augmented XTREME benchmark~\citep{hu2020xtrememassivelymultilingualmultitask} by retaining its original tasks (excluding question answering) and incorporating our CodeComplexity and MathShepherd datasets to cover a broader range of domains.

\paragraph{Model specialisation via causal ecosystem (F1 score):}

\begin{itemize}[leftmargin=15pt, itemsep=1pt, topsep=0pt]
    \item \textbf{Beaver~\citep{ji2023beavertailsimprovedsafetyalignment} -- Safety:} This classification task evaluates LLM outputs across 14 harm categories to measure content moderation capabilities.\footnote{\url{https://huggingface.co/datasets/PKU-Alignment/BeaverTails}}
 
    \item \textbf{Safe~\citep{ji2025pkusaferlhfmultilevelsafetyalignment} -- Safety:} This binary classification task is derived from RLHF preference data, where each model response is annotated as safe or unsafe.\footnote{\url{https://huggingface.co/datasets/PKU-Alignment/PKU-SafeRLHF}}
 
    \item \textbf{Aegis~\citep{ghosh2025aegis20diverseaisafety} -- Safety:} This AI content classification task spans 13 risk categories and is designed to evaluate safety filtering in generative AI systems.\footnote{\url{https://huggingface.co/datasets/nvidia/Aegis-AI-Content-Safety-Dataset-2.0}}
 
    \item \textbf{E-SNLI-VE~\citep{do2021esnlivecorrectedvisualtextualentailment} -- Image-Text English:} This visual entailment task extends E-SNLI to image--text pairs, involving the classification of whether an image entails, contradicts, or is neutral with respect to a textual hypothesis.\footnote{\url{https://huggingface.co/datasets/sedrickkeh/e-snli-ve}}
 
    \item \textbf{BoolQ-Audio -- Audio-Text English:} This audio-based Boolean question-answering task classifies spoken questions paired with a text passage as yes or no.\footnote{\url{https://huggingface.co/datasets/fixie-ai/boolq-audio}}
\end{itemize}

\subsection{General Embeddings Evaluation}

The following benchmarks evaluate general-purpose embeddings without fine-tuning the model on task-specific data. Depending on the task type, evaluation is either fully zero-shot (e.g., cosine similarity for retrieval) or uses a lightweight linear probe (e.g., logistic regression for classification). All three benchmarks are part of the MTEB ecosystem, enabling us to efficiently compare our models against thousands of baselines across a large set of tasks.\footnote{\url{https://github.com/embeddings-benchmark/mteb}}

\begin{itemize}[leftmargin=15pt, itemsep=1pt, topsep=0pt]
    \item \textbf{MTEB (English, v2)~\citep{muennighoff_mteb_2023}:} A comprehensive English text embedding benchmark derived from MTEB (English, v1). It spans seven task categories: classification, clustering, pair classification, reranking, retrieval, semantic textual similarity, and summarization, comprising a total of 41 tasks. 
    
    \item \textbf{MTEB (Multilingual, v2)~\citep{enevoldsen2025mmtebmassivemultilingualtext}} A large-scale multilingual text embedding benchmark covering 250+ languages, curated from the full MMTEB collection~\citep{muennighoff_mteb_2023} via inter-task correlation-based downsampling to reduce computational cost while preserving model rankings. Tasks span eight categories: classification, clustering, pair classification, reranking, retrieval, semantic textual similarity, bitext mining, and summarization.
 
    \item \textbf{MIEB (lite)~\citep{xiao2025miebmassiveimageembedding}} A lightweight image embedding benchmark covering 51 tasks across 10 task types, designed as a cost-efficient version of MIEB(Multilingual) while maintaining relative model rankings. Task types include clustering, few-shot linear probing, zero-shot classification, retrieval (image-to-image, text-to-image, and cross-modal), document understanding, visual STS, compositionality, and interleaved embedding evaluation.
 
    \item \textbf{MAEB (beta)~\citep{assadi2026maebmassiveaudioembedding}} An audio embedding benchmark with 30 tasks spanning audio-only and audio-text cross-modal evaluation in 100+ languages, derived from a larger 98-task collection (MAEB+). Tasks span seven types: classification~(10), retrieval~(9), clustering~(3), pair classification~(3), multilabel classification~(2), zero-shot classification~(2), and reranking~(1).
\end{itemize}

\subsection{Aggregating Performance Across Tasks}
To enable a fair comparison across tasks with heterogeneous metrics and
scales, we report an \emph{average normalized rank}.  For each task
$t \in \mathcal{T}$ and model $m \in \mathcal{M}$, let $v_{t,m}$ denote
the aggregate performance score.  We rescale every model to a
$[0,\,|\mathcal{M}|-1]$ interval via
\begin{equation}
  r_{t,m} \;=\; (|\mathcal{M}|-1)\,\cdot\,
    \frac{\max_{m'} v_{t,m'} \;-\; v_{t,m}}
         {\max_{m'} v_{t,m'} \;-\; \min_{m'} v_{t,m'}}\,,
\end{equation}
so that $r_{t,m}=0$ for the best-performing model on task $t$ and
$r_{t,m}=|\mathcal{M}|-1$ for the worst.  The overall rank of a model
is then the arithmetic mean across all tasks:
\begin{equation}
  \bar{r}_{m} \;=\; \frac{1}{|\mathcal{T}|}
    \sum_{t \in \mathcal{T}} r_{t,m}\,.
\end{equation}
A lower $\bar{r}_{m}$ therefore indicates a model that performs
consistently well across all evaluation tasks, regardless of the
individual metric used in each one.

\subsection{Evaluation Fine-Tuning Protocol}
\paragraph{Text fine-tuning protocol.} All models are fine-tuned under a consistent protocol with a batch size of 32. For each model--dataset pair, we select the learning rate from 10 log-spaced values between $5\times10^{-6}$ and $5\times10^{-3}$, using a 10\% warmup schedule followed by linear decay. To avoid data contamination during model selection and evaluation, we rely on existing training, validation, and test splits, or manually create them when unavailable. To accommodate architectural differences, we follow standard practice by using the final-token representation for causal models and mean pooling for bidirectional models on retrieval, sequence classification, and regression tasks.

\begin{itemize}[leftmargin=15pt, itemsep=1pt, topsep=0pt]
    \item \textbf{Retrieval:} Fine-tuning runs for 1k steps on MS MARCO, followed by zero-shot cross-domain retrieval on the remaining benchmarks.
    \item \textbf{Sequence regression:} Fine-tuning runs for 5k steps.
    \item \textbf{Sequence and token classification:} Fine-tuning runs for 10k steps.
\end{itemize}

For smaller datasets, which undergo multiple training epochs, we apply early stopping with a patience of one epoch based on validation performance during each fine-tuning run.

\paragraph{Post-merging specialisation fine-tuning.}
To evaluate the effectiveness of merging our adapted model with causal specialists, we assessed performance across several domain-specific tasks. We followed the previously evaluation setup with one exception: given the limited number of samples in these specialized benchmarks and to ensure perfectly balanced label distributions within each training set, we utilized a smaller batch size to guarantee a minimum of 500 training steps per task.

\begin{itemize}[leftmargin=15pt, itemsep=1pt, topsep=0pt]
    \item \textbf{Beaver:} Batch size: 4.
    \item \textbf{e-SNLI-VE:} Batch size: 32.
    \item \textbf{BoolQ-Audio:} Batch size: 14.
\end{itemize}

To accommodate the limited number of samples available in these benchmarks, and to highlight the accelerated convergence of the merged model compared to baseline, we reduce the batch size relative to the previous text benchmarks. This adjustment ensures a minimum of 500 training steps per benchmark.

\clearpage
\section{BidirLM-Omni Model Composition}\label{sec:omni_diagram}

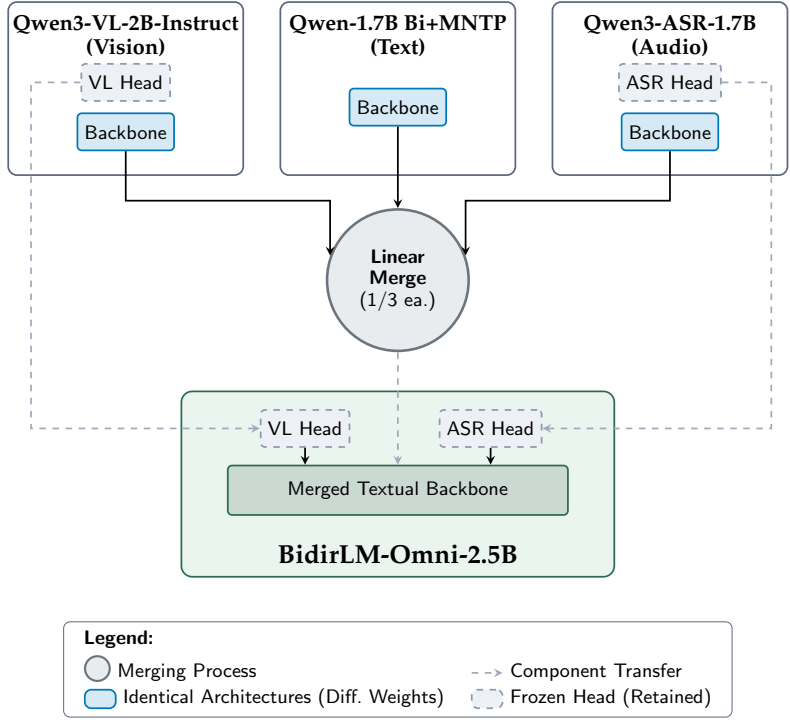
\begin{figure}[h]
\centering
\resizebox{0.75\textwidth}{!}{%
\begin{tikzpicture}[font=\small\sffamily]

    \definecolor{accentA}{HTML}{0077B6}      
    \definecolor{accentAlight}{HTML}{D6EAF5} 

    \definecolor{accentB}{HTML}{6C757D}      
    \definecolor{accentBlight}{HTML}{E9ECEF} 

    \definecolor{accentC}{HTML}{2D6A4F}      
    \definecolor{accentClight}{HTML}{E9F5EE} 

    \definecolor{frozenGray}{HTML}{8D99AE}   
    \definecolor{frozenFill}{HTML}{EDF0F4}   
    \definecolor{boxStroke}{HTML}{5C677D}     
    \definecolor{transferGray}{HTML}{A0AABB}  

    \tikzset{
        model/.style = {
            rectangle, rounded corners=4pt,
            draw=boxStroke, fill=white, thick
        },
        component/.style = {
            rectangle, rounded corners=2pt,
            minimum width=1.4cm, minimum height=0.6cm,
            text centered, draw=black, fill=white
        },
        merge/.style = {
            circle, minimum size=1.5cm, text centered,
            draw=accentB, fill=accentBlight, line width=1.2pt
        },
        arrow/.style  = {thick, ->, >=stealth},
        trainable/.style = {
            draw=accentA, fill=accentAlight, thick
        },
        frozen/.style = {
            draw=frozenGray, fill=frozenFill, dashed, thick
        }
    }


    \node (vlspecial) [model, minimum width=3.8cm, minimum height=2.8cm]
        at (-4.4, 0.4) {};
    \node [anchor=north, font=\bfseries, align=center]
        at ([yshift=-0.1cm]vlspecial.north)
        {Qwen3-VL-2B-Instruct \\ (Vision)};

    \node (mntpspecial) [model, minimum width=3.8cm, minimum height=2.8cm]
        at (0, 0.4) {};
    \node [anchor=north, font=\bfseries, align=center]
        at ([yshift=-0.1cm]mntpspecial.north)
        {Qwen-1.7B Bi+MNTP \\ (Text)};

    \node (asrspecial) [model, minimum width=3.8cm, minimum height=2.8cm]
        at (4.4, 0.4) {};
    \node [anchor=north, font=\bfseries, align=center]
        at ([yshift=-0.1cm]asrspecial.north)
        {Qwen3-ASR-1.7B \\ (Audio)};


    \node (vlhd) [component, frozen]    at (-4.4, 0.5)  {VL Head};
    \node (vlbb) [component, trainable] at (-4.4, -0.3)  {Backbone};

    \node (mntpbb) [component, trainable] at (0, 0.1) {Backbone};

    \node (asrhd) [component, frozen]    at (4.4, 0.5)  {ASR Head};
    \node (asrbb) [component, trainable] at (4.4, -0.3)  {Backbone};

    \node (mergepoint) [merge] at (0, -2.7) {
        \begin{tabular}{c}
            \textbf{Linear} \\ \textbf{Merge} \\ (1/3 ea.)
        \end{tabular}
    };

    \draw [arrow]
        (vlbb.south) -- ([yshift=-0.8cm]vlbb.south) -| (mergepoint.160);
    \draw [arrow]
        (mntpbb.south) -- (mergepoint.north);
    \draw [arrow]
        (asrbb.south) -- ([yshift=-0.8cm]asrbb.south) -| (mergepoint.20);

    \node (omnibox) [
        rectangle, rounded corners=5pt,
        draw=accentC, fill=accentClight, thick,
        minimum width=7cm, minimum height=3cm
    ] at (0, -6.0) {};

    \node[anchor=south, font=\large\bfseries, text=black]
        at ([yshift=0.1cm]omnibox.south) {BidirLM-Omni-2.5B};

    \node (omnibackbone) [component,
        draw=accentC, fill=accentC!25, thick,
        minimum width=5.5cm, minimum height=0.8cm]
        at (0, -6.1) {Merged Textual Backbone};

    \node (finalvlhd)  [component, frozen] at (-1.5, -5.1) {VL Head};
    \node (finalasrhd) [component, frozen] at ( 1.5, -5.1) {ASR Head};

    \draw [arrow, dashed, transferGray] (mergepoint.south) -- (omnibackbone.north);

    \draw [arrow, thick]
        (finalvlhd.south)  -- ([xshift=-1.5cm]omnibackbone.north);
    \draw [arrow, thick]
        (finalasrhd.south) -- ([xshift= 1.5cm]omnibackbone.north);

    \draw [arrow, dashed, transferGray]
        (vlhd.west)  -- ([xshift=-0.8cm]vlhd.west)  |- (finalvlhd.west);
    \draw [arrow, dashed, transferGray]
        (asrhd.east) -- ([xshift= 0.8cm]asrhd.east) |- (finalasrhd.east);

    \node (legend) [rectangle, rounded corners=3pt,
        draw=boxStroke, minimum width=9cm, fill=white] at (0, -9.0) {
        \begin{tabular}{ll}
            \textbf{Legend:} & \\[2pt]
            \tikz[baseline=-1ex]{
                \node[circle, minimum size=0.4cm,
                      draw=accentB, fill=accentBlight, line width=1pt] {};}
            Merging Process &
            \tikz[baseline=-0.5ex]{
                \draw[thick,->,>=stealth, transferGray, dashed](0,0)--(.5,0);}
            Component Transfer \\[2pt]
            \tikz[baseline=-0.5ex]{
                \node[rectangle, minimum width=0.5cm, minimum height=0.3cm,
                      draw=accentA, fill=accentAlight, thick]{};}
            Identical Architectures (Diff.\ Weights) &
            \tikz[baseline=-0.5ex]{
                \node[rectangle, minimum width=0.5cm, minimum height=0.3cm,
                      draw=frozenGray, fill=frozenFill, dashed, thick]{};}
            Frozen Head (Retained) \\
        \end{tabular}
    };
\end{tikzpicture}
}
\caption{The construction of BidirLM-Omni-2.5B relies on a modular composition strategy. We begin with three specialized variants sharing an identical underlying architecture: a vision model (Qwen3-VL-2B), our bidirectional text encoder (Qwen-1.7B Bi+MNTP), and an audio model (Qwen3-ASR-1.7B). First, we isolate their trainable textual backbones and perform a linear weight merge in equal proportions ($1/3$ each) to forge a unified omnimodal representation space. Second, we extract the frozen, modality-specific projection heads (visual and audio) from the specialist models and seamlessly append them to the newly merged backbone. This composition enables cross-modal routing.}
\label{fig:omni_taxonomy}
\end{figure}

\section{Additional Results}\label{sec:additional_results}
\subsection{Masked Language Objectives and Hyperparameters}

\begin{figure}[h] 
\centering 
\includegraphics[width=\textwidth]{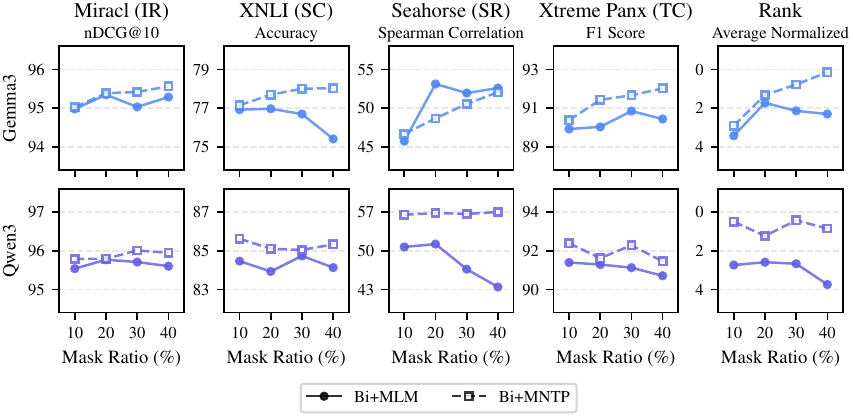}
\caption{\textbf{Performance comparison of MLM vs.\ MNTP adaptation.} The first four columns report dataset-specific scores, while the rightmost column reports the model ranking based on average normalized performance across all tasks.}
\label{fig:mlm_vs_mntp} 
\end{figure}

We evaluate the MLM and MNTP objectives for bidirectional adaptation across four masking ratios (10\%, 20\%, 30\%, and 40\%) using a 10B-token subset of FineWeb-Edu. \autoref{fig:mlm_vs_mntp} reports the downstream performance of the resulting Bi+MLM and Bi+MNTP models, along with their ranking based on average normalized scores.\footnote{To manage the extensive search space over masking ratios and learning rates, we limit XNLI and PAN-X fine-tuning to 5k steps for this comparison.}

\paragraph{MNTP outperforms MLM for model adaptation.}
As shown in \autoref{fig:mlm_vs_mntp}, Bi+MNTP consistently outperforms Bi+MLM across tasks and architectures, except on the Seahorse dataset for Gemma at 20\% and 30\%. More generally, Bi+MNTP achieves higher mean performance than Bi+MLM at every corresponding masking ratio. Furthermore, all Bi+MNTP models with masking ratios above 20\% surpass the highest average performance of any Bi+MLM variant, establishing Bi+MNTP as the stronger of the two masking objectives.

\paragraph{Optimal masking ratios are objective- and model-dependent.} 
Bi+MLM performance typically peaks at intermediate ratios (20\% and 30\%, \autoref{fig:mlm_vs_mntp}), whereas Bi+MNTP benefits from higher masking, achieving optimal average performance at 30\% for Qwen3-0.6B and 40\% for Gemma3-270M.

\subsection{Details on Model Similarities}\label{sec:details_on_merging_and_similarities}
As discussed in \autoref{sec:scaling}, the success of our weight-merging strategy relies on the observation that the adapted and causal models remain close in weight space. Here, we ground this observation in prior theoretical literature and extend our analysis to a layer-by-layer level across the various merging configurations explored in this study, providing finer-grained evidence that weight displacement remains bounded and consistent.

\paragraph{Empirical context for merging.} Prior work has shown that models fine-tuned from a shared pretrained checkpoint often remain in the same basin of the loss landscape, a property known as linear mode connectivity~\citep{frankle2020linearmodeconnectivitylottery}, enabling their convex combinations to perform well~\citep{wortsman2022model}. A complementary observation by \citet{ortizjimenez2023taskarithmetictangentspace} suggests that pretraining induces weight disentanglement, whereby distinct capabilities are encoded along approximately orthogonal directions in weight space, reducing interference when models are combined. While these results were established for models sharing the same objective and attention mechanism, we note that our setting shares a key favorable condition: all merged models derive from the identical pretrained backbone. Furthermore, our adaptation processes a small fraction of tokens relative to the original pre-training scale while maintaining next-token prediction objectives, resulting in remarkably limited weight displacement (mean cosine similarity of 0.78 for Gemma and 0.97 for Qwen).

\paragraph{Methodology.} For each model pair, we compute the layer-wise cosine similarity between corresponding weight tensors. To achieve this, we flatten and concatenate all Self-Attention and MLP parameters within a given layer into a single vector. Additionally, we disentangle these results into two specific weight groups: Self-Attention (Q, K, V, and O projections) and MLP (gate, up, and down projections).

\begin{figure}[h] 
\centering 
\includegraphics[width=\textwidth]{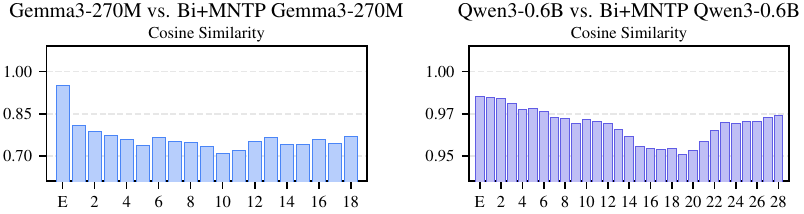}
\includegraphics[width=\textwidth]{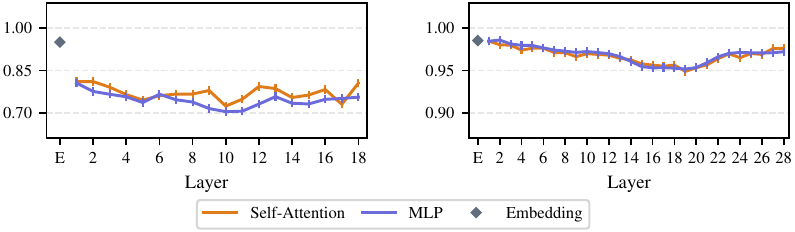}
\caption{\textbf{Per-layer cosine similarity between causal models and their Bi+MNTP-adapted encoders.} \textit{Top:} Aggregate cosine similarity per layer. \textit{Bottom:} Comparison broken down by by Self-Attention and MLP weight group.}
\label{fig:similarity_mntp} 
\end{figure}

\begin{figure}[h] 
\centering 
\includegraphics[width=\textwidth]{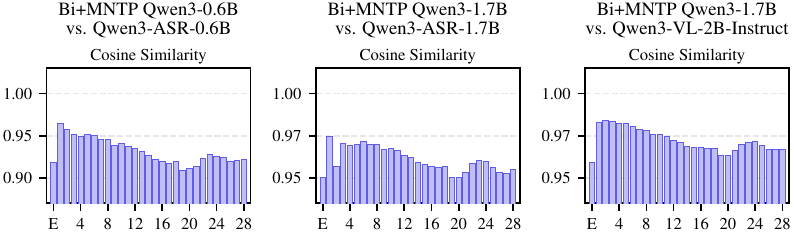}
\includegraphics[width=\textwidth]{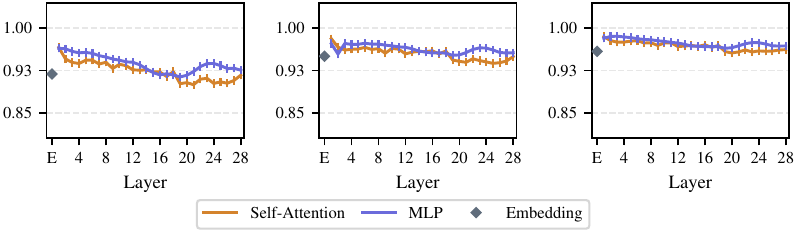}
\caption{\textbf{Per-layer cosine similarity between Bi+MNTP adapted Qwen3 models and their causal multimodal variants.} \textit{Top:} Aggregate cosine similarity per layer. \textit{Bottom:} Comparison down by Self-Attention and MLP weight group.}
\label{fig:similarity_multimodal} 
\end{figure}

\paragraph{MNTP adaptation.} \autoref{fig:similarity_mntp} reports the per-layer cosine similarity between the original causal models and resulting Bi+MNTP encoder counterparts for Gemma3-270M and Qwen3-0.6B, quantifying the weight displacement introduced by our bidirectional adaptation. The two architectures exhibit different similarity profiles: Gemma3-270M shows substantially lower overall similarity (mean cosine 0.78) compared to Qwen3-0.6B (mean cosine 0.97). When broken down by weight group, Self-Attention and MLP projections follow a similar trend, with MLP weights consistently exhibiting slightly larger deviations.

\paragraph{Multimodal variants.} \autoref{fig:similarity_multimodal} extends this analysis to the causal multimodal specialists that we used during the multimodal alignment ablation and to construct BidirLM-Omni-2.5B (\autoref{sec:alignment_and_domain_specificity}). This entails a comparison of our Bi+MNTP Qwen3-0.6B against Qwen3-ASR-0.6B (left plot), and our Bi+MNTP Qwen3-1.7B against both Qwen3-VL-2B-Instruct (which utilizes the same 1.7B text backbone) and Qwen3-ASR-1.7B. Examining these 1.7B variants (right two subplots), we observe that the similarity with the VL specialist remains slightly higher than with the ASR specialist (mean cosine 0.97 vs.\ 0.96). Furthermore, the aggregate similarity over Self-Attention and MLP projections generally decreases with depth for each model, indicating that later layers undergo the most modification. All causal models used to construct BidirLM-Omni-2.5B maintain a high overall similarity with the Bi+MNTP Qwen3-1.7B encoder ($> 0.96$).

\subsection{Performance by Merging Ratio with Causal Specialists}\label{sec:merging_ratio_causal_specialists}
\begin{figure}[tbh]
\centering
\includegraphics[width=\textwidth]{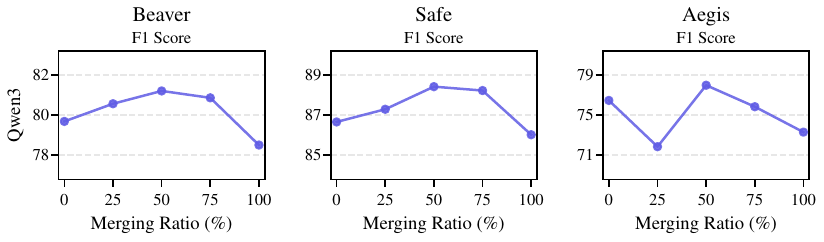}
\caption{\textbf{Model performance on safety classification benchmarks across merging ratios.} We report the resulting scores as a function of the weight allocated to the causal Qwen3Guard-Gen-0.6B model when merged with our Bi+MNTP Qwen3-0.6B encoder.}
\label{fig:guard_ratio}
\end{figure}

\begin{figure}[tbh]
\centering
\includegraphics[width=\textwidth]{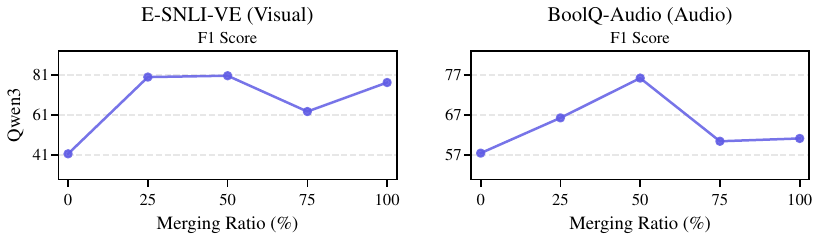}
\caption{\textbf{Model performance on multimodal classification benchmarks across merging ratios.} We report the resulting scores as a function of the weight allocated to the causal Qwen3-ASR-0.6B and Qwen3-VL-2B-Instruct models when merged with our Bi+MNTP Qwen3-0.6B and Qwen3-1.7B encoders.}
\label{fig:multimodal_ratio}
\end{figure}

Following the merging ratio analysis conducted in \autoref{fig:merging_ratio} for catastrophic forgetting mitigation, we investigate the merging behavior of encoder specialization when leveraging causal specialists for domain (\autoref{fig:guard_ratio}) and multimodal (\autoref{fig:multimodal_ratio}) adaptation.

\paragraph{An equal merging ratio emerges as a robust baseline.} Consistent with our strategy for mitigating catastrophic forgetting, an equal 50\% split consistently yields the highest performance by efficiently weighing the encoder's bidirectional capabilities against the specialized knowledge of the causal models. Therefore, we advise practitioners to adopt a 0.5 interpolation weight as a strong default, exploring nearby values to extract peak performance given that the merging process is computationally training-free.

\paragraph{Merging with non-shared modalities is ratio-sensitive.} While merging models within a common modality such as text yields robust performance even at intermediate ratios like 25\% or 75\%, merging across distinct modalities results in significant performance drops at these same unbalanced values. This indicates that as the discrepancy between the base models' domains or modalities increases, overall performance becomes highly sensitive to the interpolation weight, reinforcing a balanced 50\% split as the most reliable default choice.

\subsection{Effect of Merging on Omnimodal Performance}\label{sec:effects_of_merging_omni}
\begin{figure}[tbh]
\centering
\includegraphics[width=0.9\textwidth]{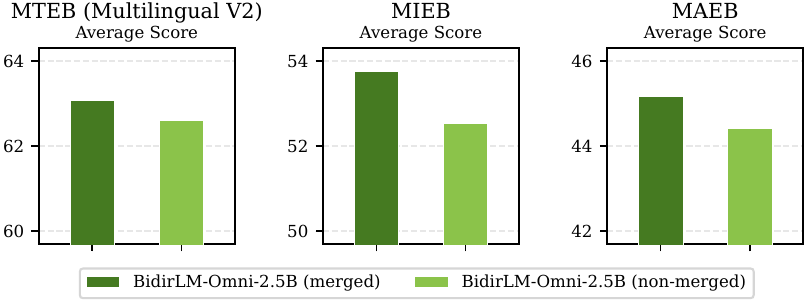}
\caption{\textbf{Average score per benchmark: BidirLM-Omni-2.5B (merged) vs.\ BidirLM-Omni-2.5B (non-merged).} Average score across individual tasks. Comparison following the contrastive training phase between the merged BidirLM-Omni variant and the non-merged baseline on: MTEB (Multilingual V2), MIEB (lite), and MAEB (beta).}
\label{fig:bar_benchmark_avg}
\end{figure}

To evaluate the impact of our merging strategy when creating BidirLM-Omni, we compare two variants: BidirLM-Omni (where each of the three specialized causal backbones contributes equally), and a non-merged baseline relying exclusively on the Bi+MNTP weights without integrating the common weights shared with the ASR and VL specialists (only concatenating their frozen multimodal heads on top). As shown in \autoref{fig:bar_benchmark_avg}, merging appears as a key performance factor, with higher scores achieved by the merged variant across all three benchmarks.

\subsection{Detailed Results Across Models and Benchmarks}

\autoref{tab:task_type_scores} reports per-task-type scores on MTEB (Multilingual V2) for our four text-only BidirLM encoders alongside the omnimodal BidirLM-Omni-2.5B. \autoref{tab:omni_task_type_mieb} and \autoref{tab:omni_task_type_maeb} further detail BidirLM-Omni-2.5B performance across MIEB (lite) and MAEB (beta) benchmarks.

\begin{table}[H]
\centering
\resizebox{\textwidth}{!}{%
\begin{tabular}{lccccccccccc}
\toprule
\textbf{MTEB (Multilingual V2)} & \shortstack{Bitext\\Mining} & Class. & Clust. & \shortstack{Instr.\\Rerank.} & \shortstack{ML\\Class.} & \shortstack{Pair\\Class.} & Rerank. & Retr. & STS & \shortstack{Mean\\(Task)} & \shortstack{Mean\\(TaskType)} \\
\midrule
BidirLM-270M & 59.5 & 58.8 & 45.9 & -2.1 & 19.7 & 76.7 & 56.8 & 48.7 & 67.7 & 55.5 & 48.0 \\
BidirLM-0.6B & 66.7 & 62.2 & 49.6 & 0.8 & 23.8 & 78.4 & 58.1 & 56.5 & 70.8 & 59.6 & 51.9 \\
BidirLM-1B & 71.8 & \textbf{65.9} & 50.3 & -0.4 & 25.1 & 79.8 & 58.6 & 56.5 & 74.6 & 62.1 & 53.6 \\
BidirLM-1.7B & \textbf{72.2} & 65.7 & 51.5 & 0.5 & \textbf{26.6} & 80.2 & 62.1 & \textbf{59.9} & 74.2 & 62.9 & 54.8 \\
BidirLM-Omni-2.5B & \textbf{72.2} & 65.5 & \textbf{51.6} & \textbf{0.9} & \textbf{26.6} & \textbf{80.7} & \textbf{63.4} & 59.4 & \textbf{75.7} & \textbf{63.1} & \textbf{55.1} \\
\bottomrule
\end{tabular}
}
\caption{Performance per task type on MTEB (Multilingual V2). Best score per column is \textbf{bolded}. Class.: classification, Clust.: clustering, Instr.\ Rerank.: instruction reranking, ML Class.: multilabel classification, Pair Class.: pair classification, Rerank.: reranking, Retr.: retrieval.}
\label{tab:task_type_scores}
\end{table}
\begin{table}[H]
\centering
\resizebox{\textwidth}{!}{%
\begin{tabular}{lccccccccc}
\toprule
\textbf{MIEB (lite)} & Retr. & Comp. & \shortstack{Doc.\\Underst.} & \shortstack{Img.\\Class.} & \shortstack{Img.\\Clust.} & \shortstack{Vision\\QA} & \shortstack{ZS\\Class.} & \shortstack{Mean\\(Task)} & \shortstack{Mean\\(TaskType)} \\
\midrule
BidirLM-Omni-2.5B & 28.8 & 46.0 & 76.9 & 73.6 & 61.1 & 48.6 & 48.1 & 58.1 & 54.7 \\
\bottomrule
\end{tabular}
}
\caption{Performance per task type on MIEB (lite) for BidirLM-Omni-2.5B. Comp.: compositionality, Vision QA: vision-centric QA, ZS Class.: zero-shot classification.}
\label{tab:omni_task_type_mieb}
\end{table}
\begin{table}[H]
\centering
\resizebox{\textwidth}{!}{%
\begin{tabular}{lccccccccc}
\toprule
\textbf{MAEB (beta)} & \shortstack{Any2Any\\Retr.} & \shortstack{Audio\\Class.} & \shortstack{Audio\\Clust.} & \shortstack{Audio\\ML Class.} & \shortstack{Audio\\Pair Class.} & \shortstack{Audio\\Rerank.} & \shortstack{Audio\\ZS Class.} & \shortstack{Mean\\(Task)} & \shortstack{Mean\\(TaskType)} \\
\midrule
BidirLM-Omni-2.5B & 32.8 & 58.2 & 5.5 & 32.3 & 66.4 & 74.8 & 55.3 & 45.2 & 46.5 \\
\bottomrule
\end{tabular}
}
\caption{Performance per task type on MAEB (beta) for BidirLM-Omni-2.5B. Any2Any Retr.: any-to-any retrieval, Audio ML Class.: audio multilabel classification, Audio ZS Class.: audio zero-shot classification.}
\label{tab:omni_task_type_maeb}
\end{table}

\subsection{MTEB, MIEB, and MAEB (2026-03-30 Snapshot).}

\begin{table}[H]
\centering
\resizebox{\textwidth}{!}{%
\begin{tabular}{lrccccccccccccc}
\toprule
\textbf{MTEB (Multilingual V2)} & \textbf{Params} & \shortstack{Mean\\Rank} & \shortstack{Zero\\Shot} & \shortstack{Bitext\\Mining} & Class. & Clust. & \shortstack{Instr.\\Rerank.} & \shortstack{ML\\Class.} & \shortstack{Pair\\Class.} & Rerank. & Retr. & STS & \shortstack{Mean\\(TaskType)} & \shortstack{Mean\\(Task)} \\
\midrule
KaLM-Gemma3-12B & 11.8B & 1 & 73\% & \textbf{83.8} & \textbf{77.9} & \textbf{55.8} & \textbf{5.5} & \textbf{33.0} & 84.7 & \textbf{67.3} & \textbf{75.7} & \textbf{79.0} & \textbf{62.5} & \textbf{72.3} \\
$\cdots$ &  &  &  &  &  &  &  &  &  &  &  &  &  &  \\
\textbf{BidirLM-Omni-2.5B} & 2.5B & 17 & 100\% & 72.2 & 65.5 & 51.6 & 0.9 & 26.6 & 80.7 & 63.4 & 59.4 & 75.7 & 55.1 & 63.1 \\
\textbf{BidirLM-1.7B} & 1.7B & 18 & 100\% & 72.2 & 65.7 & 51.5 & 0.5 & 26.6 & 80.2 & 62.1 & 59.9 & 74.2 & 54.8 & 62.9 \\
$\cdots$ &  &  &  &  &  &  &  &  &  &  &  &  &  &  \\
GTE-Qwen2-7B & 7.1B & 22 & -- & 73.9 & 61.5 & 52.8 & 4.9 & 25.5 & \textbf{85.1} & 65.5 & 60.1 & 74.0 & 55.9 & 62.5 \\
$\cdots$ &  &  &  &  &  &  &  &  &  &  &  &  &  &  \\
\textbf{BidirLM-1B} & 1.0B & 24 & 100\% & 71.8 & 65.9 & 50.3 & -0.4 & 25.1 & 79.8 & 58.6 & 56.5 & 74.6 & 53.6 & 62.1 \\
$\cdots$ &  &  &  &  &  &  &  &  &  &  &  &  &  &  \\
\textbf{BidirLM-0.6B} & 596M & 36 & 100\% & 66.7 & 62.2 & 49.6 & 0.8 & 23.8 & 78.4 & 58.1 & 56.5 & 70.8 & 51.9 & 59.6 \\
$\cdots$ &  &  &  &  &  &  &  &  &  &  &  &  &  &  \\
BGE-M3 & 568M & 38 & 98\% & 79.1 & 60.4 & 40.9 & -3.1 & 20.1 & 80.8 & 62.8 & 54.6 & 74.1 & 52.2 & 59.6 \\
$\cdots$ &  &  &  &  &  &  &  &  &  &  &  &  &  &  \\
GTE-Qwen2-1.5B & 1.5B & 40 & -- & 62.5 & 58.3 & 52.0 & 0.7 & 24.0 & 81.6 & 62.6 & 60.8 & 71.6 & 52.7 & 59.5 \\
$\cdots$ &  &  &  &  &  &  &  &  &  &  &  &  &  &  \\
KaLM-mini & 494M & 49 & 92\% & 64.8 & 57.6 & 45.6 & -1.5 & 20.7 & 77.7 & 60.6 & 54.2 & 70.8 & 50.0 & 57.0 \\
$\cdots$ &  &  &  &  &  &  &  &  &  &  &  &  &  &  \\
Stella-1.5B & 1.5B & 54 & 90\% & 58.6 & 56.7 & 49.7 & 0.2 & 21.8 & 78.5 & 61.4 & 52.8 & 69.9 & 50.0 & 56.5 \\
$\cdots$ &  &  &  &  &  &  &  &  &  &  &  &  &  &  \\
\textbf{BidirLM-270M} & 268M & 58 & 100\% & 59.5 & 58.8 & 45.9 & -2.1 & 19.7 & 76.7 & 56.8 & 48.7 & 67.7 & 48.0 & 55.5 \\
$\cdots$ &  &  &  &  &  &  &  &  &  &  &  &  &  &  \\
Nomic-v1 & 137M & 87 & 96\% & 28.4 & 49.4 & 40.8 & -3.2 & 17.8 & 71.7 & 46.0 & 37.0 & 64.2 & 39.1 & 45.7 \\
\bottomrule
\end{tabular}
}
\caption{Per-task-type performance on MTEB (Multilingual V2) for our models (bold) and open-data baselines, ranked by Mean (Task). $\cdots$ denotes a jump in leaderboard entries. Zero-shot ratios from the \href{https://huggingface.co/spaces/mteb/leaderboard}{MTEB leaderboard}. Best per column in \textbf{bold}.}
\label{tab:task_type_scores_all}
\end{table}

\begin{table}[tbh]
\centering
\resizebox{\textwidth}{!}{%
\begin{tabular}{lrccccccccccc}
\toprule
\textbf{MIEB (lite)} & \textbf{Params} & \shortstack{Mean\\Rank} & \shortstack{Zero\\Shot} & Retr. & Comp. & \shortstack{Doc.\\Underst.} & \shortstack{Img.\\Class.} & \shortstack{Img.\\Clust.} & \shortstack{Vision\\QA} & \shortstack{ZS\\Class.} & \shortstack{Mean\\(TaskType)} & \shortstack{Mean\\(Task)} \\
\midrule
\textbf{BidirLM-Omni-2.5B} & 2.5B & 2 & 100\% & 28.8 & \textbf{46.0} & \textbf{76.9} & 73.6 & 61.1 & 48.6 & 48.1 & \textbf{54.7} & \textbf{58.1} \\
$\cdots$ &  &  &  &  &  &  &  &  &  &  &  &  \\
E5-V & 8.4B & 6 & 100\% & 26.9 & 39.4 & 56.0 & 70.6 & 51.7 & 52.6 & 36.2 & 47.6 & 51.9 \\
$\cdots$ &  &  &  &  &  &  &  &  &  &  &  &  \\
Nemotron-Omni-3B & 4.7B & 8 & -- & 31.8 & 41.0 & 46.6 & 68.3 & 75.6 & 47.9 & 43.7 & 50.7 & 51.4 \\
CLIP-bigG & 2.5B & 9 & 100\% & 34.2 & 35.0 & 35.5 & 77.8 & 80.8 & 43.0 & 72.4 & 54.1 & 51.3 \\
$\cdots$ &  &  &  &  &  &  &  &  &  &  &  &  \\
EVA02-bigE+ & 5.0B & 11 & 100\% & \textbf{35.2} & 38.9 & 26.2 & \textbf{80.0} & \textbf{87.3} & 38.8 & \textbf{74.0} & 54.4 & 50.8 \\
$\cdots$ &  &  &  &  &  &  &  &  &  &  &  &  \\
CLIP-L-DC & 428M & 13 & 100\% & 31.0 & 31.6 & 30.8 & 75.3 & 80.4 & 54.9 & 69.4 & 53.4 & 50.4 \\
CLIP-H & 986M & 14 & 100\% & 32.8 & 34.8 & 33.7 & 76.8 & 79.3 & 46.8 & 69.4 & 53.4 & 50.0 \\
$\cdots$ &  &  &  &  &  &  &  &  &  &  &  &  \\
CLIP-B16-DC & 150M & 22 & 100\% & 28.3 & 31.4 & 22.7 & 73.2 & 73.6 & 56.9 & 61.9 & 49.7 & 46.6 \\
$\cdots$ &  &  &  &  &  &  &  &  &  &  &  &  \\
VLM2Vec & 4.1B & 24 & 100\% & 20.9 & 30.2 & 42.8 & 64.8 & 65.4 & \textbf{65.3} & 32.1 & 45.9 & 46.0 \\
\bottomrule
\end{tabular}
}
\caption{Per-task-type performance on MIEB (lite) for our models (bold) and open-data baselines, ranked by Mean (Task). $\cdots$ denotes a jump in leaderboard entries.}
\label{tab:omni_task_type_mieb_all}
\end{table}

\begin{table}[H]
\centering
\resizebox{\textwidth}{!}{%
\begin{tabular}{lrccccccccccc}
\toprule
\textbf{MAEB (beta)} & \textbf{Params} & \shortstack{Mean\\Rank} & \shortstack{Zero\\Shot} & \shortstack{Any2Any\\Retr.} & \shortstack{Audio\\Class.} & \shortstack{Audio\\Clust.} & \shortstack{Audio\\ML Class.} & \shortstack{Audio\\Pair Class.} & \shortstack{Audio\\Rerank.} & \shortstack{Audio\\ZS Class.} & \shortstack{Mean\\(TaskType)} & \shortstack{Mean\\(Task)} \\
\midrule
LCO-Omni-7B & 8.9B & 1 & -- & \textbf{55.2} & 57.1 & 1.7 & \textbf{38.4} & \textbf{67.3} & 78.7 & 64.5 & \textbf{51.9} & \textbf{52.0} \\
LCO-Omni-3B & 4.7B & 2 & -- & 54.5 & 55.4 & 1.3 & 36.8 & 66.7 & 75.4 & 62.2 & 50.3 & 50.7 \\
\textbf{BidirLM-Omni-2.5B} & 2.5B & 3 & 100\% & 32.8 & \textbf{58.2} & 5.5 & 32.3 & 66.4 & 74.8 & 55.3 & 46.5 & 45.2 \\
$\cdots$ &  &  &  &  &  &  &  &  &  &  &  &  \\
Nemotron-Omni-3B & 4.7B & 5 & -- & 38.8 & 47.6 & 1.2 & 21.1 & 66.1 & \textbf{81.5} & \textbf{69.8} & 46.6 & 43.1 \\
$\cdots$ &  &  &  &  &  &  &  &  &  &  &  &  \\
MS-CLAP-23 & 160M & 9 & 100\% & 18.7 & 45.0 & 15.2 & 7.0 & 53.6 & 75.4 & 12.6 & 32.5 & 31.3 \\
CLAP-fused & 154M & 10 & 100\% & 17.6 & 44.4 & \textbf{22.7} & 4.5 & 52.0 & 61.3 & 13.2 & 30.8 & 30.8 \\
CLAP-unfused & 153M & 11 & 100\% & 18.1 & 45.2 & 12.6 & 4.3 & 52.6 & 66.5 & 11.3 & 30.1 & 30.3 \\
MS-CLAP-22 & 196M & 12 & 100\% & 21.7 & 38.2 & 19.9 & 7.2 & 51.7 & 62.9 & 12.1 & 30.5 & 29.8 \\
$\cdots$ &  &  &  &  &  &  &  &  &  &  &  &  \\
SpeechT5 & 298M & 15 & 100\% & 7.9 & 40.7 & 1.1 & 8.5 & 57.9 & 56.5 & 15.9 & 26.9 & 25.3 \\
\bottomrule
\end{tabular}
}
\caption{Per-task-type performance on MAEB (beta) for our models (bold) and open-data baselines, ranked by Mean (Task). $\cdots$ denotes a jump in leaderboard entries.}
\label{tab:omni_task_type_maeb_all}
\end{table}

\end{document}